\documentclass[preprint,11pt]{elsarticle}
\usepackage[english]{babel}
\usepackage{amssymb,amsmath,graphicx}
\usepackage{pifont}
\usepackage{amsthm}
\usepackage{float}
\usepackage{afterpage}
\usepackage{color}
\usepackage{enumitem}
\usepackage{setspace}
\usepackage[table]{xcolor}
\usepackage{multirow}
\usepackage{microtype}
\usepackage[status=draft]{fixme}
\usepackage{subcaption}
\usepackage{adjustbox}
\usepackage{longtable}
\usepackage{rotating}
\usepackage{siunitx} 
\usepackage{xcolor,colortbl}
\definecolor{Gray}{gray}{0.8}

\usepackage{url}
\usepackage{booktabs}

\usepackage[ruled,linesnumbered,noend]{algorithm2e}
\usepackage{tikz}
\usetikzlibrary{arrows}
\usetikzlibrary{decorations.markings}
\usepackage{pgfplots, pgfplotstable} 

\newcommand{\cmark}{\ding{51}}%
\newcolumntype{H}{>{\setbox0=\hbox\bgroup}c<{\egroup}@{}}

\newcommand{\myred}[1]{#1}

\allowdisplaybreaks
\selectfont

\begin{document}
\begin{center}
 
\begin{LARGE}
PILS: Exploring high-order neighborhoods by\vspace*{0.1cm}\linebreak pattern mining and injection
\end{LARGE}
 
\vspace*{01cm}
 
\textbf{Florian Arnold$^a$, \'{I}talo Santana$^b$, Kenneth S\"{o}rensen$^a$, Thibaut Vidal$^{b*}$} \\
 $^a$ University of Antwerp, Department of Engineering Management, Belgium\\
\{florian.arnold,kenneth.sorensen\}@uantwerpen.be\\
 $^b$Departamento de Inform\'{a}tica, Pontif\'{i}cia Universidade Cat\'{o}lica do Rio de Janeiro (PUC-Rio) \\
 \{isantana,vidalt\}@inf.puc-rio.br \\
 
\vspace*{0.8cm}

\end{center}
 
\noindent
\textbf{Abstract.} We introduce \emph{pattern injection local search} (PILS), an optimization strategy that uses pattern mining to explore high-order local-search neighborhoods, and illustrate its application on the vehicle routing problem. PILS operates by storing a limited number of frequent patterns from elite solutions. During the local search, each pattern is used to define one move in which 1)~incompatible edges are disconnected, 2)~the edges defined by the pattern are reconnected, and 3)~the remaining solution fragments are optimally reconnected. Each such move is accepted only in case of solution improvement. As visible in our experiments, this strategy results in a new paradigm of local search, which complements and enhances classical search approaches in a controllable amount of computational time. We demonstrate that PILS identifies useful high-order moves (e.g., \textsc{9-opt} and \textsc{10-opt}) which would otherwise not be found by enumeration, and that it significantly improves the performance of state-of-the-art population-based and neighborhood-centered metaheuristics.\\
 
 
\noindent
$^*$ Corresponding author

\thispagestyle{empty}
\pagenumbering{arabic}

\section{Introduction} 
\label{sec:introduction}

Since the \emph{``no free lunch''} theorem \citep{Wolpert1997}, it is known that no method can ---~on average~--- perform better than any other method on all possible problems. For combinatorial optimization problems, this theorem implies that specialized algorithms need to be designed that attempt to exploit the specific \emph{structure} of the problem to be solved. Recent research \citep[e.g.,][]{Arnold2019} has demonstrated that discovering the structural properties of high-quality solutions, i.e., what differentiates high-quality from low-quality solutions, can be instrumental in developing state-of-the-art heuristics. In this paper, we investigate whether \emph{pattern mining}, i.e., the discovery of frequently used patterns or structures in high-quality solutions, can similarly improve the performance of a heuristic optimization algorithm. 

Pattern mining is a well-established technique to detect correlations and substructures in datasets. It is traditionally used in market-data analysis to identify sets of products that are frequently acquired together, and many other applications exist, such as DNA analysis and fraud detection~\citep{Aggarwal2014}. In all of these cases, the extraction of patterns reveals insightful associations and can guide strategic decisions.

We focus this study on the \emph{capacitated vehicle routing problem} (CVRP). This problem belongs to the class of optimization problems known as \emph{vehicle routing problems}. These problems seek to find least-cost delivery routes to visit a geographically dispersed customer set, therefore generalizing the classical traveling salesman problem (TSP) with multiple vehicles and other side constraints \citep{Toth2014,Vidal2019}. Almost all vehicle routing problems are NP-hard as an extension of the classical TSP, but most are notoriously more difficult to solve in practice. Despite 60 years of research and published research papers numbering in the thousands, the best exact algorithms for vehicle routing problems remain unable to consistently solve instances with around 300 customers in a reasonable amount of computation time \citep{Costa2019,Pecin2017}. In contrast, the largest TSP instance to be solved to \emph{proven optimality} to date has a whopping 85,900 cities \citep{Applegate2009}. Due to both their computational difficulty and practical interest, vehicle routing problems have therefore emerged as one of the most important benchmarks for (meta)heuristics, designed to produce high-quality approximate solutions in a controlled time.

It is well known that high-quality solutions of a vehicle routing problem tend to be structurally close to the global optima, with which they share a large number of common edges \citep{Boese1995}. Moreover, during a typical search, several sequences of consecutive visits regularly re-appear in high-quality solutions. A few studies have attempted to exploit such \emph{patterns} heuristically, either by guiding the search towards frequently occurring customer sequences or by building new initial solutions from them as a starting point for the local search operators. However, state-of-the-art heuristics for vehicle routing problems generally rely on efficient local search operators to a far greater extent than on iterative solution construction procedures. We therefore posit that a careful adaptation of the local search components using a set of high-quality patterns (i.e., customer sequences that frequently occur in high-quality solutions) could be a promising avenue in the design of high-quality of heuristics for vehicle routing problems. This research path, however, remains mostly unexplored.

To fill this gap, we introduce a technique to effectively exploit discovered patterns in a local search heuristic. We have called this technique \emph{pattern injection local search} (PILS). PILS is a generic move generator that efficiently finds high-order moves (i.e., moves in which more than two visits are affected simultaneously) based on patterns frequently occurring in high-quality solutions.

In a nutshell, PILS consists of two algorithmic steps: pattern collection and pattern injection. \emph{Pattern collection} is the process of collecting patterns (i.e., sequences of consecutive visits) that frequently occur in high-quality solutions. Then, a subset of the most frequent patterns can be introduced in an incumbent solution in a three-step process called \emph{pattern injection}. (1)~Incompatible edges (i.e., edges adjacent to nodes in the pattern, but not occurring in the pattern itself) are disconnected. (2)~The edges defined by the pattern are reconnected. This yields a set of disconnected route fragments, which are (3)~optimally reconnected. In PILS, a pattern injection move is only accepted if it improves the incumbent solution. 

The ability of PILS to find high-order pattern injection moves can be easily used to complement other local searches and is independent of the metaheuristic paradigm used (e.g., population- or trajectory-based methods). We demonstrate this generality by applying PILS in the framework of two state-of-the-art metaheuristics for the CVRP: the hybrid genetic search of \citet{Vidal2014} and the guided local search of \mbox{\citet{Arnold2019}}.
In summary, the contributions of this work are threefold:
\begin{enumerate}[nosep]
 \item[1)] We introduce a new optimization technique called \emph{pattern injection local search} (PILS) that can be used to generate high-order moves by introducing patterns frequently discovered in high-quality solutions in an incumbent solution. We discuss the major design decisions and implementation strategies related to this new technique. To the best of our knowledge, this paper presents the first attempt to use pattern mining to generate and enumerate specialized large neighborhoods.
\item[2)] As part of the PILS approach, we describe a simple algorithm to optimally reconnect the route fragments that occur during the pattern injection phase.
\item[3)] Finally, we conduct extensive experiments to measure the effectiveness of PILS using two state-of-the-art metaheuristics for the CVRP. We also evaluate how pattern frequency and quality are correlated, and measure the sensitivity of the approach to the number of selected patterns and the number of pattern-insertion attempts, thereby providing a deep analysis of the role of pattern mining in local search-based metaheuristics.
\end{enumerate}

The remainder of this paper is organized as follows. Section~\ref{sec:literature-review} reviews the related literature. Section~\ref{sec:proposed-methodology} describes the PILS methodology, while Section~\ref{sec:pils-application} discusses the integration of PILS within two state-of-the-art metaheuristics for the CVRP. Section~\ref{sec:experiments} presents our computational experiments, and Section~\ref{sec:conclusion} concludes.

\section{Literature review}
\label{sec:literature-review}

\paragraph{Pattern mining and metaheuristics}
If we (informally) define a pattern as \emph{a set of solution characteristics}, then pattern extraction and exploitation is, at least indirectly, a founding principle of most modern metaheuristics. According to \citet{Holland1992}, the success of crossover-based genetic algorithms is largely because they promote the survival and propagation of high-quality building blocks. Similarly, path relinking algorithms \citep{Resende2010} iteratively guide the search towards the characteristics of an elite solution, while ant colony optimization (ACO) \citep{Dorigo2019} learns and reinforces promising decisions. This modus operandi comes from the fact that, for most combinatorial optimization problems of interest, high-quality solutions are structurally close to the global optimum in the solution space \citep{Boese1995}.

Other recent metaheuristics have more directly exploited pattern information, for two general purposes: (1) information exchange and cooperation between the various operators used in the metaheuristic, and (2) to generate new initial solutions. \citet{LeBouthillier2005} rely on frequent patterns to coordinate and guide the search of several metaheuristics. Patterns are extracted from a \emph{solution warehouse} and used to temporarily fix or prohibit edges in cooperating tabu searches and genetic algorithms.
\citet{ElHachemi2015} and \citet{Lahrichi2015} have extended this methodology into an integrative cooperative search (ICS) for multi decision-attribute optimization problems, relying on structural problem decompositions and \emph{integrations} of partial elite solutions to form complete solutions.

While studies on pattern guidance remain few and far between, contributions in which patterns are exploited to generate new initial solutions are more widespread.
Adaptive memory programming (AMP -- \citep{Taillard2001}) is a methodological paradigm that represents this strategy well. It generalizes most of the classical metaheuristics (tabu search, scatter search, genetic algorithms, and ACO) within a unified framework, based on the premises that all these methods ``memorize solutions or characteristics of solutions generated during the search process'' and ``include a procedure that creates an initial solution with the information stored in memory''. \textsc{BoneRoute} \citep{Tarantilis2002} successfully applies the AMP strategy to the CVRP within a population-based approach. New partial solutions are regularly built from solution components and completed heuristically. Similarly, \citet{Santos2006} proposes a genetic algorithm, in which new solutions are generated via a multi-parent crossover or a construction procedure combining elite patterns. Set-covering-based \emph{matheuristics} \citep{Muter2010,Subramanian2013} also regularly combine solution elements (e.g., routes, bins, clusters) into complete solutions using integer programming solvers. Several studies have also focused on identifying frequent sequences or visits to construct new initial solutions for the TSP, leading to methods known as backbone search \citep{Kilby2005,Schneider2003}, tabu search with vocabulary building \citep{Glover1997}, and fixed set search \citep{Jovanovic2019}.

It is tempting to combine multiple promising solution fragments into new solutions. However, search methods based on this principle face a major problem: even though several \emph{individual} decisions may be found in a large number of high-quality solutions, \emph{combinations} of these decisions may not. For example, even though there may exist edges that appear in a large number of high-quality solutions of a vehicle routing problem, this does not in any way guarantee that any combination of these promising edges can be used to form a high-quality feasible solution. In other words, the pattern built from promising decisions is generally not \emph{supported} in any high-quality solution. For this reason, \cite{Barbalho2013,Ribeiro2006} and other related studies opted to use a single ---~large and supported~--- pattern during each~solution~construction.

To summarize, previous studies have, either directly or indirectly, exploited pattern mining to enhance metaheuristics. Coined \emph{parts}, \emph{fragments} or \emph{backbones}, these patterns capture frequent structures from elite solutions. To the best of our knowledge, patterns have been mainly used to guide the search and drive solution construction rather than local improvement, a surprising fact given that local searches play the most critical role in most modern metaheuristics. We therefore aim to design new strategies to exploit pattern information at the local search level, through specialized, enumerable moves whose evaluation complexity remain controllable, leading to a new local search paradigm.

\paragraph{The capacitated vehicle routing problem}
Due to its importance for transportation logistics and its rich combinatorial structure, the CVRP currently stands as one of the main benchmarks for research on combinatorial optimization algorithms. In its canonical form, it is defined on a complete graph $\mathcal{G}=(\mathcal{V},\mathcal{E})$ such that $\mathcal{V}= 0 \cup \{1,\ldots,n\}$. Vertex $0$ stands for a depot where a vehicle fleet is based, and each other vertex $i\in\{1,\dots,n\}$ represents a customer with demand~$q_i$. Each edge $(i,j)\in \mathcal{E}$ represents the possibility of traveling from $i$ to $j$ with distance cost $c_{ij} \in \mathbb{R}^+$. The goal of the CVRP is to design up to $m$ vehicle routes starting and ending at the depot, in such a way that each customer is visited once, that the total demand transported on each route does not exceed the vehicle capacity~$Q$, and that the total cost measured as the sum of the route distances is minimized~\mbox{\citep{Toth2014}}.

The CVRP is NP-hard as a generalization of the TSP. Despite the considerable progress of mathematical programming techniques for NP-hard combinatorial optimization problems, current exact methods for the CVRP can only solve instances with a few hundred customers in a reasonable amount of time \citep{Costa2019,Pecin2017}. Since this size is insufficient for recent applications, e.g., for e-commerce or mobility-on-demand, extensive research has been conducted on metaheuristics in an attempt to generate approximate solutions in a more controlled computational effort. Similarly, metaheuristics have regularly appeared in the pattern recognition and machine learning domains, for optimization tasks involving very large datasets \citep{Bahrololoum2017,Gribel2019,Han2019a,Hansen2012,Ijjina2016,Kashef2015a,Yusta2009}.

All of the classical metaheuristic paradigms have been tested on the CVRP.
In the early 2000s, state-of-the-art algorithms were primarily based on tabu search and other single-trajectory metaheuristics~\citep{Gendreau1994}. This status-quo changed with the proposal of effective hybrid genetic searches (HGS) for this problem \citep{Nagata2009,Prins2004,Vidal2012,Vidal2014}, combining the exploration abilities of crossover- and population-based search with the improvement potential of specialized local searches to achieve a fine balance between diversification and intensification \citep{Blum2003,Vidal2012a}. The algorithm of \citet{Vidal2012} has been holding the best-known results for the CVRP for nearly a decade. Recently, two other methods have achieved high-quality results for some instance classes: the adaptive large neighborhood search with \emph{slack induction} of \citet{Christiaens2018} and the knowledge-guided local search (KGLS) of \citet{Arnold2019}. As visible at \url{http://vrp.galgos.inf.puc-rio.br/index.php/en/updates}, research remains very active on the topic, and new best solutions are still regularly reported for the classical instances of \mbox{\citet{Uchoa2017}}.

\section{Pattern Injection Local Search}
\label{sec:proposed-methodology} 

Nearly all successful CVRP metaheuristics rely on some local search-based optimization component, which is iteratively applied on multiple solutions throughout the method. Given an incumbent solution $s$, a local search (LS) explores a neighborhood $\mathcal{N}(s)$ which includes all solutions reachable from $s$ by small changes, called \emph{moves}, with the goal of finding an improving neighbor which is used as a new incumbent solution. This process is repeated until reaching a \emph{local minimum} state, where no more improving neighbor exists. It can be said that LS performs a mapping of a set of initial solutions onto a set of local minima, which can be seen as a discrete analogy to gradient descent in continuous space. 

Neighborhood size is typically exponential in the number of vertices or edges which are jointly modified in a move: e.g., there exist $\Theta(k! \, n^k)$ solutions in the \textsc{$k$-opt} neighborhood, obtained by deleting $k$ edges and reconnecting the resulting solution fragments. For this reason, most CVRP metaheuristics use simple $O(n^2)$-sized neighborhoods based on single-vertex relocations (\textsc{Relocate}), pairwise exchanges (\textsc{Swap}), or replacements of two edges (\textsc{2-opt}, and \textsc{2-opt*})~\citep{Vidal2012a}. Due to their larger computational requirements, higher-order neighborhoods are only rarely considered.

This is where the proposed technique PILS provides a meaningful alternative. Instead of exhaustively exploring high-order \textsc{$k$-opt} neighborhoods, it relies on the information of frequent patterns to select and consider fewer ---~targeted~--- moves that insert a pattern in the incumbent solution and optimally reconstruct the remaining edges to avoid large disruptions. We now describe the two algorithmic steps involved in this process, pattern extraction and pattern injection, and then discuss important design choices when using PILS.

\subsection{Pattern extraction}
\label{sec:pattern_extraction}

Pattern extraction, in the case of the CVRP, consists of monitoring historical solutions to obtain the frequency of patterns, i.e., the appearance of sequences of consecutive customer visits of a certain size range $\{L_\textsc{min},\ldots,L_\textsc{max}\}$. Consider a route $\sigma = (\sigma_1,\dots,\sigma_{|\sigma|})$ starting and ending at the depot, such that $\sigma_1=0$ and $\sigma_{|\sigma|}=0$. In this route, each contiguous subsequence of $(\sigma_2,\dots,\sigma_{|\sigma|-1})$ represents a (supported) pattern, which could contain either all the visited customers or a part thereof. Route $(0,1,3,5,2,6,0)$, for example, contains two patterns of size four: $(1,3,5,2)$ and $(3,5,2,6)$. As we will conduct experiments on symmetric CVRP datasets, mirrored subsequences will be considered as identical, e.g., $(1,3,5,2)=(2,5,3,1)$. In these conditions, any route~$\sigma$ contains $\max\{0,|\sigma|-1-l\}$ patterns of size $l$, and any solution contains $O(n)$ patterns of a given size, such that pattern extraction can be done by simple inspection.
This process is described in Algorithm~\ref{alg_extraction}. The resulting patterns and their associated frequency are stored in an associative array $\mathcal{A}$.

\begin{algorithm}[htbp]
\linespread{1.2}\selectfont
 \caption{Extraction of all patterns of size $l \in \{L_\textsc{min},\ldots,L_\textsc{max}\}$ from a solution $s$}
 \label{alg_extraction}
 \For{\emph{each pattern size $l \in \{L_\textsc{min},\ldots,L_\textsc{max}\}$}}
 {
 \For{\emph{each route $\sigma$ of $s$}}
 {
 \For{$i \in \{l+1,\ldots,|\sigma|-1\}$}
 {
 $p=(\sigma_{i-l+1},\ldots, \sigma_{i})$\;
 \If{$p \in \mathcal{A}$}{
 Increment the frequency of $p$ by one unit
 }
 \Else{
 Add new pattern in $p$ in $\mathcal{A}$
 }
 }
 } 
 }
\end{algorithm}

Depending on the size of the problem instance and the number of solutions from which patterns are extracted, the number of uniquely encountered patterns in $\mathcal{A}$ can be large. Since we aim to focus on a limited subset of frequent patterns during injections, we use an additional min-heap data structure to track the $\Phi_\textsc{Freq}$ most frequent patterns of each given length~$l \in \{L_\textsc{min},\ldots,L_\textsc{max}\}$. This data structure allows $O(1)$ access to the root to verify if the least frequent element of the heap needs to be replaced, and $O(\text{log}(|\Phi_\textsc{Freq}|))$ updates whenever the frequency of an element of the heap is incremented. It also allows the method to efficiently iterate over the most frequent patterns during the pattern injection phase.

\begin{figure}[htbp]
 \centering
 \begin{subfigure}[b]{0.38\textwidth}
 \centering
 \includegraphics[width=1\textwidth]{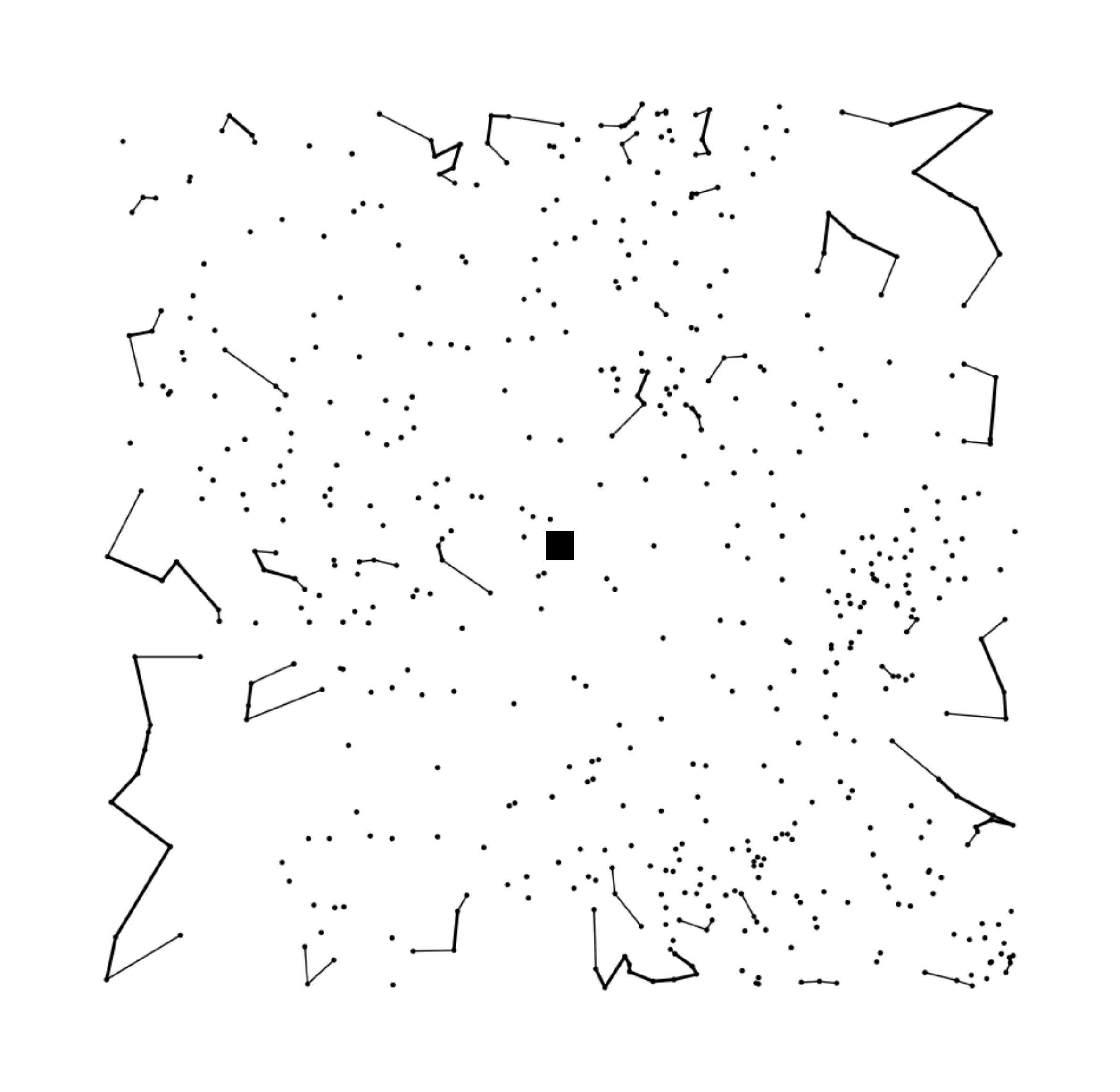}
 \caption{\small $\Phi_\textsc{Freq} = 100$, $l = 3$} 
 \end{subfigure}
 \begin{subfigure}[b]{0.38\textwidth}
 \centering
 \includegraphics[width=1\textwidth]{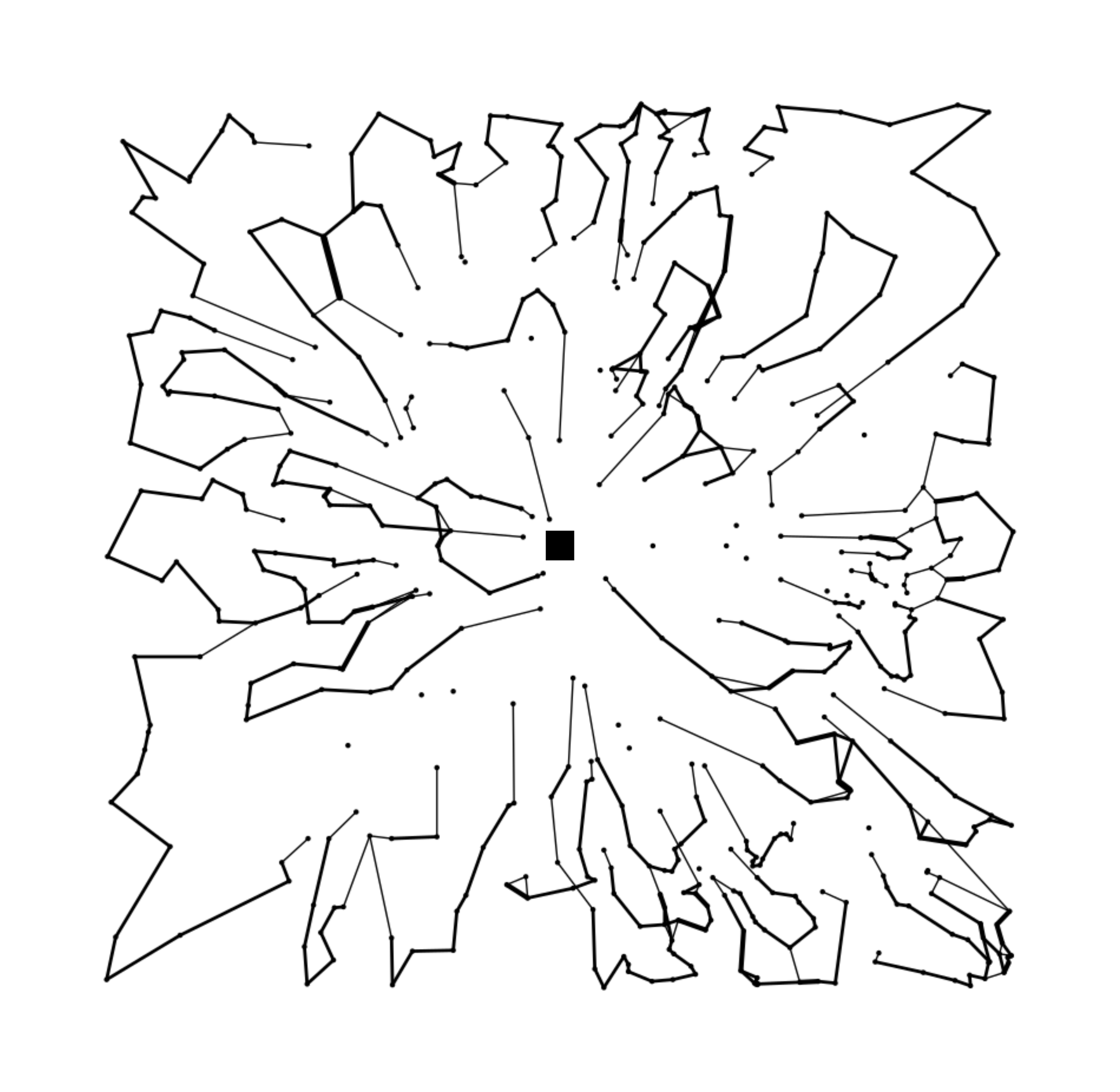}
 \caption{\small $\Phi_\textsc{Freq} = 500$, $l = 3$} 
 \end{subfigure}
 \begin{subfigure}[b]{0.38\textwidth}
 \centering
 \includegraphics[width=1\textwidth]{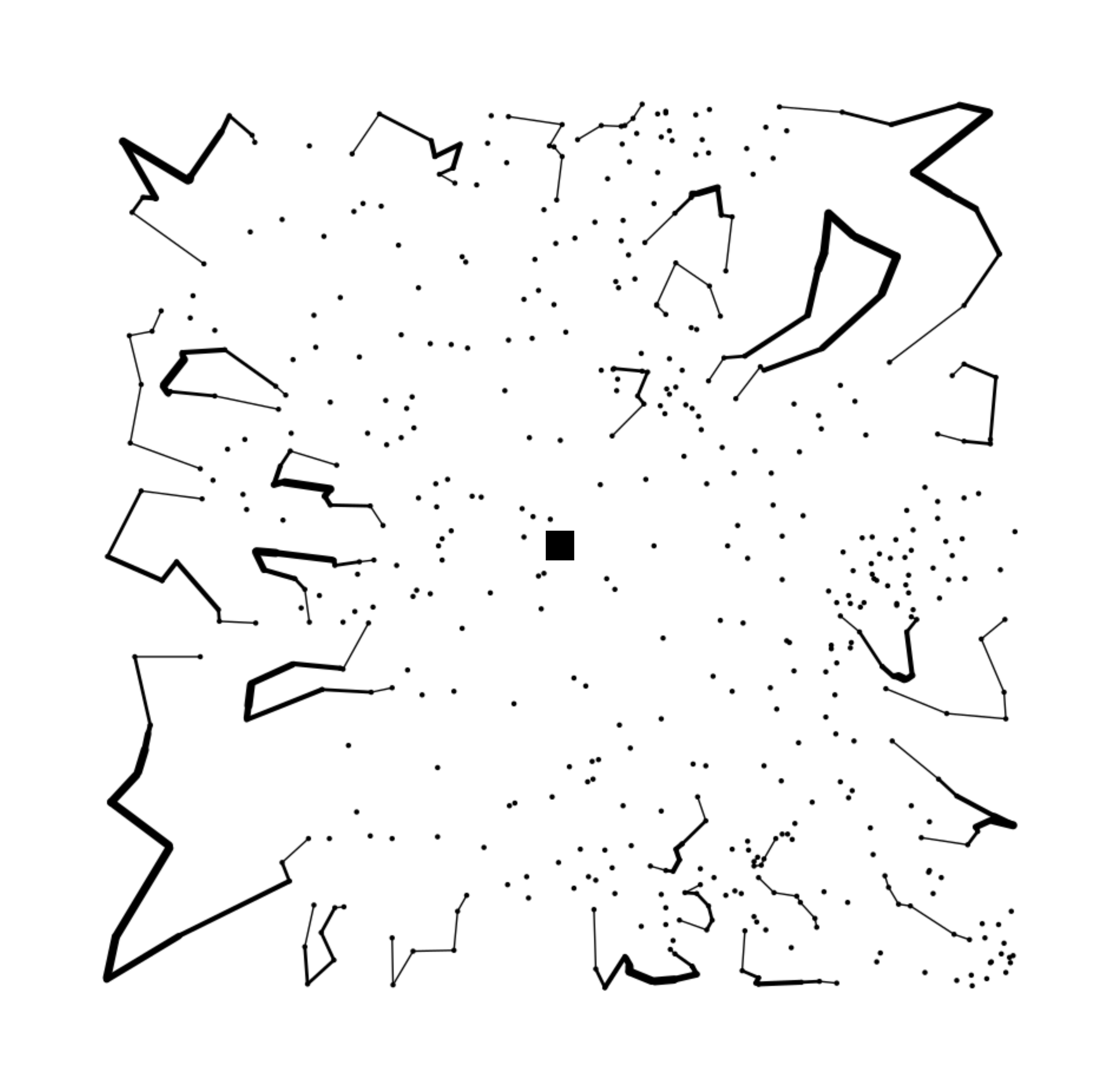}
 \caption{\small $\Phi_\textsc{Freq} = 100$, $l = 6$} 
 \end{subfigure}
 \begin{subfigure}[b]{0.38\textwidth}
 \centering
 \includegraphics[width=1\textwidth]{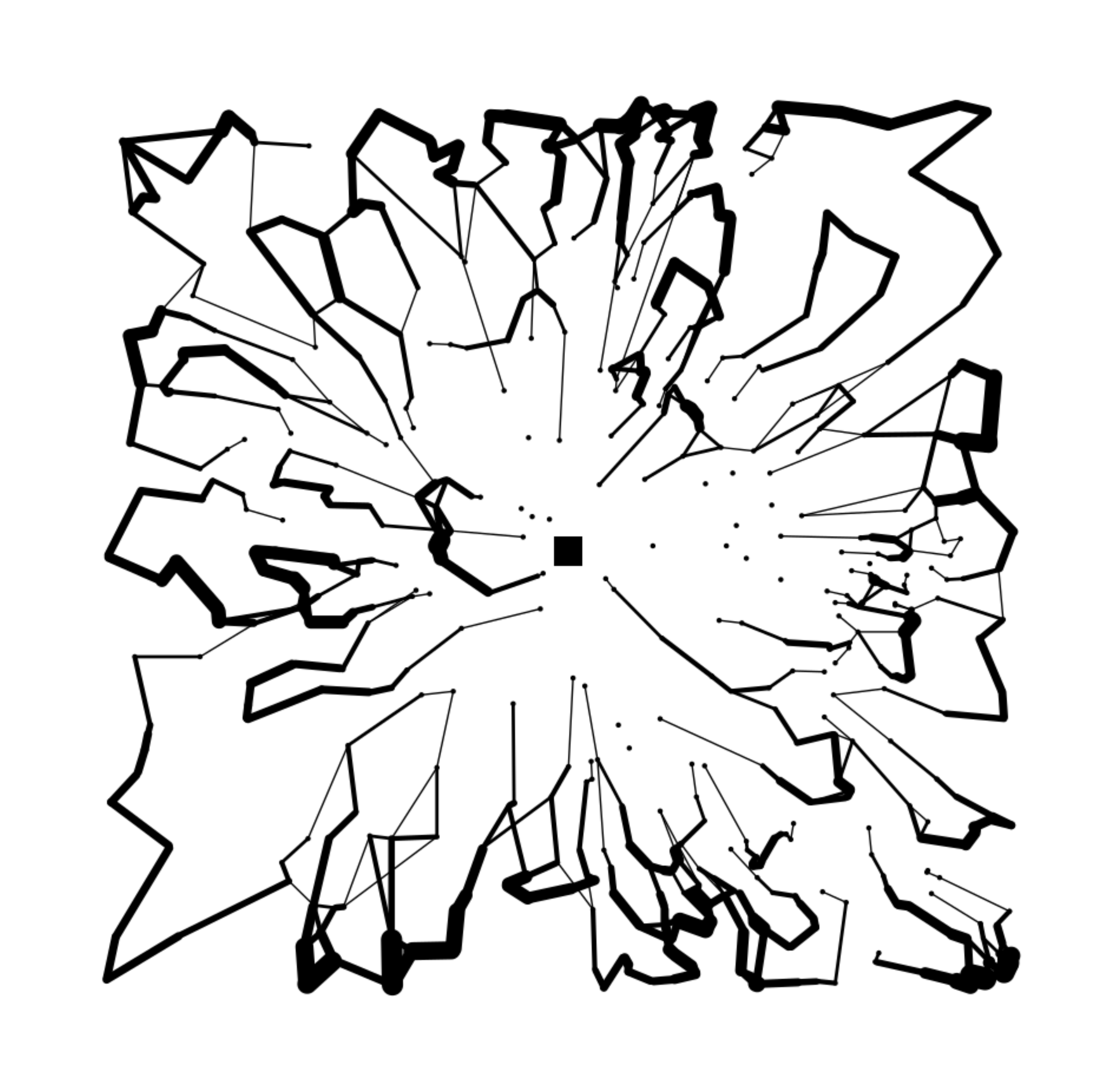}
 \caption{\small $\Phi_\textsc{Freq} = 500$, $l = 6$} 
 \end{subfigure}
 \caption{Most frequent patterns for a CVRP instance with 560 delivery locations}
 \label{interaction_extracted_patterns}
\end{figure}

Figure \ref{interaction_extracted_patterns} illustrates the $100$ and $500$ most frequent patterns of size $3$ and $6$ for a CVRP instance with $560$ delivery locations (X-n561-k42). The depot is located at the center of the figure. The thickness of the edges is proportional to their occurrence frequency in the patterns. As visible in this figure, small patterns are often contained in larger patterns. Moreover, frequent patterns usually involve customers that are more distant from the depot, since the number of relevant visit sequences for such customers tends to be smaller.

\subsection{Pattern injection}
\label{sec:pattern_injection}

During the injection phase, frequent patterns are tentatively inserted in the incumbent solution to define high-order local search moves. These moves are accepted in case of improvement. A pattern $p$ is injected into a solution by connecting the vertices of $p$ and rigorously removing all other interfering edges. This leads to a set of route fragments that need to be reconnected to obtain a feasible solution. Given an incumbent solution $s$ and a subset $\mathcal{P}$ of frequent patterns, neighborhood $\mathcal{N}_\textsc{pils}(s,\mathcal{P})$ is therefore defined as the set of all solutions obtained by:
\begin{enumerate}[nosep]
\item[1)] selecting a pattern $p \in \mathcal{P}$ which does not currently appear in $s$,
\item[2)] disconnecting in $s$ all edges adjacent to the vertices of $p$, 
\item[3)] inserting edges to form the pattern $p$, and
\item[4)] optimally inserting additional edges to obtain a complete solution.
\end{enumerate}

\begin{figure}[htbp]
 \centering
 \vspace*{0.4cm}
 \includegraphics[width=0.88\textwidth]{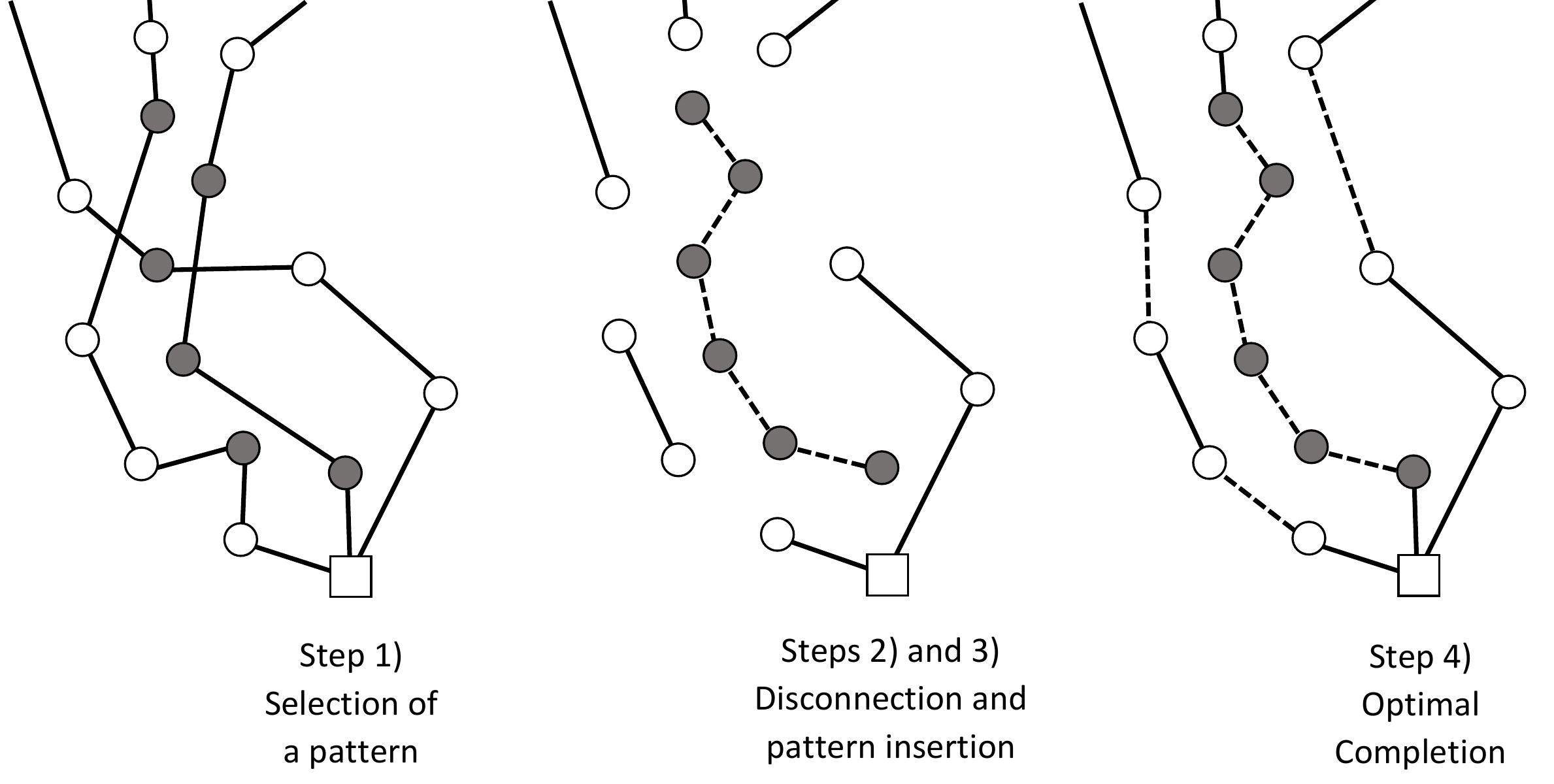}
 \caption{Illustration of the pattern injection process}
 \label{injection_graph}
\end{figure}

Figure~\ref{injection_graph} illustrates the injection process.
In this example, a pattern of size six has been selected. The pattern injection step leads to a new solution in which eight edges (represented with dashed lines) have been replaced, i.e., a \textsc{8-opt} move.

Let $\mathcal{R}_\textsc{init}$ represent the set of routes containing at least one customer of $p$. After Step 3, the routes in $\mathcal{R}_\textsc{init}$ have been partitioned into fragments which can be classified into three sets: $\mathcal{R}_\textsc{beg}$ contains all fragments that start with a depot, $\mathcal{R}_\textsc{mid}$ contains all fragments without a depot (including pattern $p$), and $\mathcal{R}_\textsc{end}$ includes all fragments that end with a depot. Note that some route fragments in $\mathcal{R}_\textsc{beg} \cup \mathcal{R}_\textsc{end}$ may contain only the depot.

During Step 4, these fragments will be optimally reconnected into a set of feasible routes via Algorithm~\ref{alg_reconnection}. Since we work with symmetric CVRP instances, fragments can potentially be reversed during this process. An efficient algorithm for this step is critical since the number of possible recombinations grows exponentially with the number of fragments. Therefore, our algorithm relies on \emph{pruning} techniques to detect and abort non-improving route combinations as early as possible in the recursions. $\mathcal{R}_\textsc{best}$ is a global variable that represents the best set of reconnected routes found so far. It is initially set to $\mathcal{R}_\textsc{best} = \mathcal{R}_\textsc{init}$. Variable $\mathcal{R}$ represents the complete routes which are currently being built. As depicted in Line 1 of Algorithm~\ref{alg_reconnection}, the route fragments are recursively concatenated (operation~$\oplus$) as long as the collective cost of the current fragments in $\mathcal{R}_\textsc{beg}$, $\mathcal{R}_\textsc{mid}$, $\mathcal{R}_\textsc{end}$, and $\mathcal{R}$ remains smaller than that of $\mathcal{R}_\textsc{best}$. In this procedure, the cost of a set of routes (or route fragments) is given by the sum of the cost of its elements: $C(\mathcal{R}) = \sum_{\sigma \in \mathcal{R}} C(\sigma)$.

\begin{algorithm}[htbp]
\linespread{1.2}\selectfont

 \caption{$\textsc{Best-Reconnect}(\mathcal{R}_\textsc{beg}, \mathcal{R}_\textsc{mid}, \mathcal{R}_\textsc{end},\mathcal{R})$.}
 \label{alg_reconnection}
 \If{$C(\mathcal{R}_\textsc{beg} \cup \mathcal{R}_\textsc{mid} \cup \mathcal{R}_\textsc{end} \cup \mathcal{R}) < C(\mathcal{R}_\textsc{best}) $}
 {
 \If{$|\mathcal{R}_\textsc{beg}| = 0$}
 {
 $\mathcal{R}_\textsc{best} = \mathcal{R}$
 }
 \Else 
 {
 Select $\sigma_\textsc{beg} \in \mathcal{R}_\textsc{beg}$\\
 \For{$\sigma_\textsc{mid} \in \mathcal{R}_\textsc{mid}$}
 {

 $\textsc{Best-Reconnect}(\mathcal{R}_\textsc{beg}-\{\sigma_\textsc{beg}\}\cup \{\sigma_\textsc{beg} \oplus \sigma_\textsc{mid}\}, \mathcal{R}_\textsc{mid}-\{\sigma_\textsc{mid}\}, \mathcal{R}_\textsc{end},\mathcal{R})$\\

 $\textsc{Best-Reconnect}(\mathcal{R}_\textsc{beg}-\{\sigma_\textsc{beg}\}\cup\{\sigma_\textsc{beg} \oplus \textsc{Rev}(\sigma_\textsc{mid})\}, \mathcal{R}_\textsc{mid}-\{\sigma_\textsc{mid}\}, \mathcal{R}_\textsc{end},\mathcal{R})$\\

 }
 \If{$|\mathcal{R}_\textsc{beg}| \neq 1$ \emph{or} $|\mathcal{R}_\textsc{mid}|=0$}
 {
 \For{$\sigma_\textsc{end} \in \mathcal{R}_\textsc{end}$}
 {
 $\textsc{Best-Reconnect}(\mathcal{R}_\textsc{beg}-\{\sigma_\textsc{beg}\}, \mathcal{R}_\textsc{mid}, \mathcal{R}_\textsc{end}-\{\sigma_\textsc{end}\},\mathcal{R} \cup \{\sigma_\textsc{beg} \oplus \sigma_\textsc{end}\})$\\
 }
 }
 }
 }
\end{algorithm}

In each recursion, one single fragment $\sigma_\textsc{beg}$ of $\mathcal{R}_\textsc{beg}$ is tentatively concatenated with each fragment $\sigma \in \mathcal{R}_\textsc{mid} \cup \mathcal{R}_\textsc{end}$ and each reversed fragment $\textsc{Rev}(\sigma)$ for $\sigma \in \mathcal{R}_\textsc{mid}$ (Line~8). Each such concatenation leads to a recursive call. During all recursive calls, we invariably have that \mbox{$|\mathcal{R}_\textsc{beg}| = |\mathcal{R}_\textsc{end}| = |\mathcal{R}_\textsc{init}| - |\mathcal{R}|$}. This value represents the number of routes that still need to be built. Whenever only one route remains, we do not permit a connection to the last fragment of $\mathcal{R}_\textsc{end}$ unless all fragments in $\mathcal{R}_\textsc{med}$ have been exhausted (Line~9). When this last condition occurs, the base case (Line~2) is finally attained and a possible reconnection has been obtained. At this point, due to the filtering condition, its cost is known to be smaller than the best known, and therefore $\mathcal{R}$ can be updated (Line~3).

This algorithm can be used to penalize or prohibit capacity-constraint violations in the routes. In the former case, a linear penalty term is added in the cost evaluation functions. In the latter case, the recursion is stopped in case of infeasibility (same as setting an infinite penalty).
To efficiently perform the concatenations and cost evaluations, each fragment $\sigma$ within the algorithm is characterized by a total demand $Q(\sigma)$ and distance $D(\sigma)$. Whenever a concatenation operation~$\oplus$ between fragments is performed, the associated capacities and distances are derived for the new fragment. Equations~(\ref{eq:capacity}--\ref{eq:distance}) compute these values by induction on the concatenation operation in $O(1)$ time:
\begin{align}
&Q(\sigma_1 \oplus \sigma_2 ) = Q(\sigma_1) + Q(\sigma_2)&\label{eq:capacity}\\
&D(\sigma_1 \oplus \sigma_2 ) = D(\sigma_1) + d_{\sigma_{1}(|\sigma_{1}|) \sigma_{2}(1) } + D(\sigma_2).&\label{eq:distance}
\end{align}
Based on this information, the cost of a route or fragment of route can be evaluated as:
\begin{equation}
C(\sigma) = \omega^Q \max\{Q(\sigma)-Q,0\} + D(\sigma),
\end{equation}
where $\omega^Q$ represents the penalty factor for each unit load excess over the vehicle capacity $Q$. The best solution reconnection found is applied in case of improvement over $\mathcal{R}_\textsc{init}$, otherwise, the solution remains unchanged.

\subsection{Design choices and parameters}
\label{sec:design_PILS}

Three main decisions need to be taken when applying PILS within a metaheuristic: (1)~which solutions are used for pattern extraction, (2)~which patterns are injected, and (3)~which solutions are submitted to pattern injection.

The success of PILS primarily depends on its ability to extract diverse patterns from high-quality solutions. Indeed, a pool of diverse but low-quality patterns (similar to those found in random solutions) would mainly lead to random moves. In contrast, an overly-restricted pattern set would lead to few possible injections and to excessive guidance towards the same solution characteristics and a resulting loss of diversity.
To achieve a meaningful trade-off between these two extremes, our method uses pattern extraction with a fixed probability of $P_\textsc{ex}$ on each local minimum produced by the metaheuristic. 
This design choice, i.e., only extracting patterns from local minima, guarantees a good correlation between pattern frequency and pattern quality while at the same time maintaining diversity (see Section \ref{sec:freq-quality}). 
Moreover, probability $P_\textsc{ex}$ drives the computational effort allocated to pattern extraction without changing the characteristics of the extracted patterns.

Regarding the selection of patterns for injection, we observe again that a good performance comes from a trade-off between quality and diversity. In particular, injecting all $\Phi_\textsc{Freq}$ frequent patterns
would either result in a low diversity whenever $\Phi_\textsc{Freq}$ is small, or into a large computational effort whenever $\Phi_\textsc{Freq}$ is larger. To achieve a better compromise between diversity and computational effort, a subset $\Phi_\textsc{Size} < \Phi_\textsc{Freq}$ of frequent patterns is selected for injection. These patterns are randomly selected from the heap to form the set of candidate patterns~$\mathcal{P}$. PILS then performs a single search loop over the entire neighborhood $\mathcal{N}_\textsc{pils}(\mathcal{P},\mathcal{S})$ (iterating over all patterns in $\mathcal{P}$) and directly applies every improving move. 
Finally, we opted to apply PILS immediately before the local search phases in the respective metaheuristics to maximize its impact on the search trajectory.

\section{Application of PILS in two CVRP metaheuristics}
\label{sec:pils-application} 

To demonstrate the robustness and generality of PILS, we study its application within two state-of-the-art metaheuristics for the CVRP: the hybrid genetic search (HGS) of \citet{Vidal2012}, and the knowledge-guided local search (KGLS) of \citet{Arnold2019}. While both metaheuristics produce high-quality solutions on classical test instances, they are also structurally very different. HGS evolves a diversified pool of solutions using recombination and local search operations, whereas KGLS improves a single incumbent solution in successive steps via a sophisticated local search based on ejection chains. The following paragraphs discuss the main components of these methods and their extension with~PILS.

Proposed in \citet{Vidal2012}, HGS (also called HGSADC or UHGS) combines the exploration capabilities of evolutionary algorithms, the improvement capabilities of local searches, and advanced population-diversity management schemes into a very effective solution method for vehicle routing problems. This metaheuristic uses the classical order crossover (OX) and giant-tour solution representation of \citep{Prins2004} to generate new solutions that are improved by local search with the classical \textsc{Relocate}, \textsc{Swap}, \textsc{2-opt} and \textsc{2-opt*} neighborhoods. Population diversity is preserved during the search via an active population management and biased fitness function which favors diverse and high-quality individuals, as well as active diversification phases which consists of reintroducing new initial solutions in the population.
Due to its simplicity and generality, HGS emerged as the first algorithm able to produce state-of-the-art results for over sixty vehicle routing problem variants and other permutation-based problems, finding the best-known results for thousands of classical benchmark instances \citep{Toth2014}.
To adapt this method, we simply include the pattern extraction step of PILS with probability $P_\textsc{ex} = 10\%$ after the classical local search, and the pattern injection step before it. The structure of the resulting algorithm is displayed in Algorithm~\ref{alg_HGSADC}.

\begin{algorithm}
\linespread{1.2}\selectfont
 \caption{HGS with PILS}
 \label{alg_HGSADC}
 Generate initial population\\
 \While{\emph{CPU time $< T_\textsc{max}$}}
 {
 Select $P_1$ and $P_2$ in the population\\
 Generate offspring $C$ by crossover of $P_1$ and $P_2$\\
 Apply \textsc{Pattern Injection} on $C$\\
 Apply \textsc{Local Search} on $C$ (using \textsc{Relocate}, \textsc{Swap}, \textsc{2-opt} and \textsc{2-opt*})\\
 \lIf{\emph{$C$ is infeasible}}{Repair $C$}
 With probability $P_\textsc{ex}$, apply \textsc{Pattern Extraction} on $C$\\
 Select survivors whenever maximum population size is attained\\
 \lIf{\emph{$It_\textsc{div}$ iterations without improvement}}{Diversify population}
 }
\end{algorithm}

KGLS \citep{Arnold2019} is a local-search based metaheuristic with a single solution trajectory that embeds sophisticated and complementary local search operators into a guided-local search framework \citep{voudouris2003guided}. The local search relies on \textsc{Cross-Exchange} and \textsc{Relation-Chains} operators for inter-route improvement, as well as Lin-Kernighan heuristic \citep{lin1973effective} for intra-route solution improvement. From an initial solution obtained from a construction procedure, KGLS iteratively identifies and penalizes a subset of undesirable edges with large width and length and applies the local search algorithm, which will be therefore guided towards new solutions. Moreover, very sophisticated neighborhood restrictions and data structures contribute to enhance the effectiveness of the local search, resulting in a scalable algorithm that effectively solves very large CVRP instances \cite{Arnold20192}. To apply PILS within KGLS, we simply include the pattern injection function immediately before each local search phase with $P_\textsc{ex} = 100\%$ (since KGLS generates fewer local minima than HGS), and the pattern extraction function afterward, as described in Algorithm \ref{alg_KGLS}.

\begin{algorithm}[htbp]
\linespread{1.2}\selectfont
 \caption{KGLS with PILS.}
 \label{alg_KGLS}
 Construct an initial solution $S$\\
 Apply \textsc{Local Search} on $S$\\
 \While{\emph{CPU time $< T_\textsc{max}$}}
 {
 Penalize undesirable edges in $S$\\
 Apply \textsc{Pattern injection} on $S$\\
 Apply \textsc{Local Search} on $S$ (using \textsc{Cross-Exchange}, \textsc{Relocation-Chains} and Lin-Kernighan algorithm)\\
 Apply \textsc{Pattern extraction} on $S$
 }
\end{algorithm}

\section{Computational Experiments}
\label{sec:experiments}

The goal of our computational experiments is threefold. A first experiment aims as setting the parameters of PILS and estimate the impact of PILS' main design decisions and parameters on its performance. In a second experiment we examine the main corollary underpinning the PILS heuristic: that pattern frequency and quality are correlated, i.e., that high-quality patterns more frequently appear in high-quality solutions. A final experiment attempts to evaluate the impact of different instance characteristics on the performance of PILS and the estimate the benefits of PILS when integrated into state-of-the-art metaheuristics in general. Each of these analyses will be covered in a dedicated subsection.

All experiments are executed on the classical CVRP datasets of \citet{Uchoa2017}, containing 100 benchmark instances with $100$ to $1000$ customer requests, different depot configurations (R$=$random, C$=$centered, E$=$eccentric), customer distributions (R$=$uniform, C$=$clustered, RC$=$mixed), and different average route length (short routes with large customer demands relative to the vehicle capacity, or longer routes with small customer demands relative to the vehicle capacity). We integrate PILS into the HGS and KGLS metaheuristics as specified in Section~\ref{sec:pils-application}, leading to method variants which will be called HGS-PILS and KGLS-PILS. HGS is coded in C++ and compiled with GCC 7.2.0, whereas KGLS uses Java 9.0.4.
In all experiments, these methods are run on a single core of a Xeon X5675 3.07\,GHz with 16\,GB of RAM. 

\subsection{Impact of PILS parameters}
\label{sec:impact_parameters}

The functioning of PILS is determined by three main parameters: $\Phi_\textsc{Freq}$, $\Phi_\textsc{Size}$, and $L_\textsc{max}$. Parameter $\Phi_\textsc{Freq}$ is the number of most-frequent patterns that are monitored in the heap and therefore drives the diversity of the search. Larger values allow tentative insertions of a more diverse set of patterns, whereas smaller values guide the search towards fewer \emph{elite} patterns. Parameters $\Phi_\textsc{Size}$ and $L_\textsc{max}$ control the number of patterns of each size that are tentatively injected in each search phase and the maximum pattern size respectively. These two parameters establish a trade-off between computational effort and solution improvement potential. Large patterns, in particular, can lead to higher-order PILS moves that are difficult to find otherwise, but their injections require reconnecting a larger number of solution fragments, leading to larger recursion depths in Algorithm~\ref{alg_reconnection}.

We first evaluate the sensitivity of HGS-PILS and KGLS-PILS to changes in these parameters. Starting from a baseline configuration in which $\Phi_\textsc{Freq}=5n$, $\Phi_\textsc{Size}=n$ and $L_\textsc{max}=5$ obtained from an initial calibration, we vary each parameter in turn (a so-called ``one factor at a time'' or OFAT analysis) to measure its impact on HGS-PILS and KGLS-PILS. For each configuration and problem instance, we execute the two methods five times with different random seeds and initial solutions, setting a CPU time limit linearly proportional to the number of customers, using 240 seconds for each 100 customers. The average results of these configurations over all instances and runs are reported in Table~\ref{tbl:sensitivity}. The left part of the table describes the investigated parameter configurations, whereas the right part of the table reports, for each of the two methods, the average solution quality and the fraction of the total computing time (in percent) used by PILS (extraction and injection). The quality of the solution is reported as a gap to the optimal or best-known solution value for this instance, computed for each instance as $\textsc{Gap(\%)} = 100(z - z_\textsc{bks})/z_\textsc{bks}$, where $z$ is the solution value obtained by the method and $z_\textsc{bks}$ represents the optimal or best known solution value (BKS) for this instance in the literature. 
Good solutions therefore correspond to gap values that are close to zero.

\begin{table}[htbp]
\caption{Parameter sensitivity analysis}
\label{tbl:sensitivity}
\centering
\renewcommand{\arraystretch}{1.2}
\scalebox{0.95}
{
\begin{tabular}{cccccccccc} 
\toprule
 && & & & \multicolumn{2}{c}{HGS} & & \multicolumn{2}{c}{KGLS}\\\cmidrule{6-7}\cmidrule{9-10}
PILS&$\Phi_\textsc{Freq}$ & $\Phi_\textsc{Size}$ &$L_\textsc{max}$ & & \multicolumn{1}{c}{Gap(\%)} & \multicolumn{1}{c}{T$_\textsc{PILS}$(\%)} & & \multicolumn{1}{c}{Gap(\%)} & \multicolumn{1}{c}{T$_\textsc{PILS}$(\%)} \\\toprule
\rowcolor{Gray} ON & 5n & n & 5& & 0.242 & 45.86 & & 0.520 & 7.28 \\
\rowcolor{Gray} OFF & -- & -- & -- & & 0.273 & -- & & 0.555 & -- \\
\midrule
ON & 5n & n & 3 & & 0.267 & 23.66 && 0.546 & 3.01 \\
ON & 5n & n & 4 & & 0.248 & 35.18 && 0.536 & 4.93 \\
ON & 5n & n & 6 & & 0.261& 55.89 & & 0.525 & 9.78 \\
ON & 5n & n & 7 & & 0.284& 64.53 & & 0.525 & 12.45 \\
\midrule
ON & 5n & 0.2n & 5 & & 0.257 & 15.47 & & 0.534 & 2.29 \\
ON & 5n & 0.5n & 5 & & 0.246 & 30.45 & & 0.522 & 4.25 \\
ON & 5n & 1.5n & 5 & & 0.258 & 55.56 & & 0.537 & 10.03 \\
ON & 5n & 2n & 5 & & 0.258 & 62.23& & 0.534 & 12.61\\
\midrule
ON & 2n & n & 5 & & 0.274 & 44.44 & & 0.532 & 6.09 \\
ON & 3n & n & 5 & & 0.255 & 44.98 & & 0.529 & 6.63 \\
ON & 10n & n & 5 & & 0.262 & 47.22 & & 0.535 & 8.13 \\
ON & 20n & n & 5 & &0.270& 48.56 & & 0.527 & 8.96 \\
\bottomrule
\end{tabular}
}
\end{table}

Remarkably, the wide majority of the considered HGS-PILS configurations as well as all the KGLS-PILS configurations lead to performance improvements over the baseline configuration in which PILS is deactivated (second line in Table~\ref{tbl:sensitivity}). Some search parameters such as $L_\textsc{max}$ and $\Phi_\textsc{Freq}$ have a larger influence of the method performance, whereas the value of $\Phi_\textsc{Size}$ has less impact.

Inserting too large patterns with $L_\textsc{max}= 7$ or beyond leads to a diminished final solution quality, since the large number of solution fragments needing reconnection significantly increases the share of time spent in Algorithm~\ref{alg_reconnection}. Similarly, inserting only small patterns with  $L_\textsc{max}= 3$ does not allow to fully exploit PILS search capabilities.

The influence of $\Phi_\textsc{Freq}$ on search performance is also visible. Small values of this parameter should be avoided, as they lead to reduced search diversity and worse performance, whereas large values distribute the computational effort of PILS over too many (possibly less frequent) patterns.

Parameter $\Phi_\textsc{Size}$ directly drives the number of tentative insertions and the share of time spent in PILS before reaching the termination criterion. Interestingly, HGS-PILS configurations with $\Phi_\textsc{Size} = 1.5$ or $2.0$ spend more than 60\% of their total CPU time in PILS, but still perform better than a simple deactivation of PILS. This means that the time spent within PILS is at least as meaningful for the search success as the time spent in a conventional local search, which represents most of the remaining computational effort. 

Finally, we note that PILS represents a smaller proportion of KGLS-PILS computational effort (13\% at most) than that of HGS-PILS. This is due to the fact that KGLS relies on a trajectory-based search with more sophisticated and time-consuming local-search operators than HGS (\textsc{Cross-Exchange}, \textsc{Relocation Chains}, and the Lin-Kernighan heuristic), such that the share of time spent in PILS is naturally smaller. Despite this behavioral difference, it is notable that our baseline configuration, in which $\Phi_\textsc{Freq}=5n$, $\Phi_\textsc{Size}=n$, and $L_\textsc{max}=5$, represents a good choice for both algorithms. We therefore opted to maintain this configuration for the remainder of the paper.

\subsection{Pattern frequency versus solution quality}
\label{sec:freq-quality}

\begin{figure}[htbp]
\centering
\vspace*{0.15cm}
\includegraphics[scale=0.4]{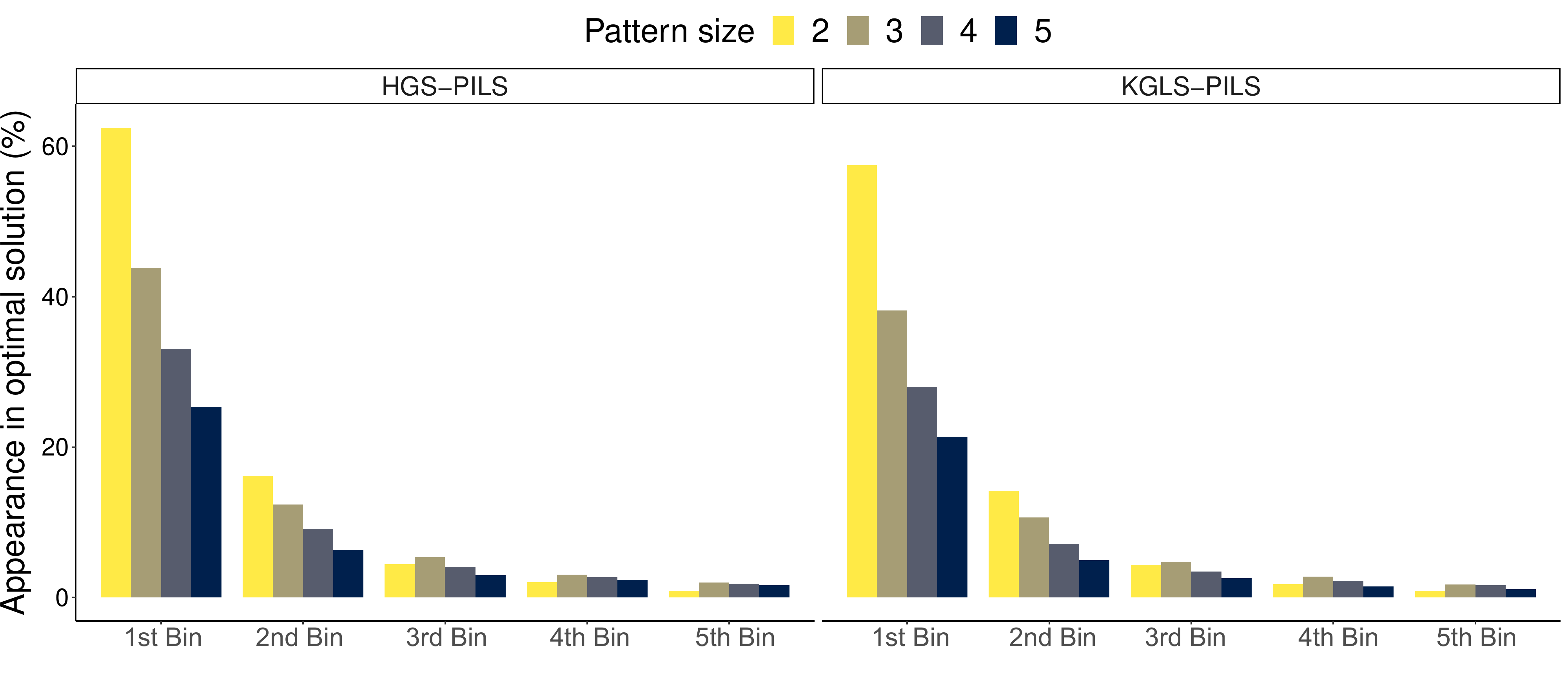}
\caption{Pattern frequency and appearance in optimal solutions}
\label{fig:opt_likelihood}
\end{figure}

Our second set of experiments investigates the relation between the frequency of the patterns and the quality of the solutions in which they appear.
For this experiment, we use a smaller subset of 10 instances with 200 to 300 delivery locations for which optimal solutions are known. We run our baseline configuration and interrupt the search after $20\%$ of the CPU time to analyze the pattern pool at an early stage of the search. For each pattern size, we sort the resulting patterns from most frequent to least frequent and distribute them into equal-sized bins containing $n$ patterns each. Finally, we calculate in each bin the fraction of patterns that appear in the optimal solution. The result of this analysis is displayed in Figure~\ref{fig:opt_likelihood}.

The results of this analysis demonstrate, for both HGS-PILS and KGLS-PILS, that the most frequent patterns (left) have a much higher probability to be part of optimal solutions than the less frequent ones (right). Moreover, the probability to belong to the optimal solution decreases when the pattern size $l$ grows, highlighting that larger optimal patterns are generally more difficult to identify. This behavior was expected since long patterns are much more informative on the structure of optimal CVRP solutions and therefore likely to be more difficult to identify.

\begin{figure}[htbp]
\centering
\vspace*{0.15cm}
\includegraphics[scale=0.4]{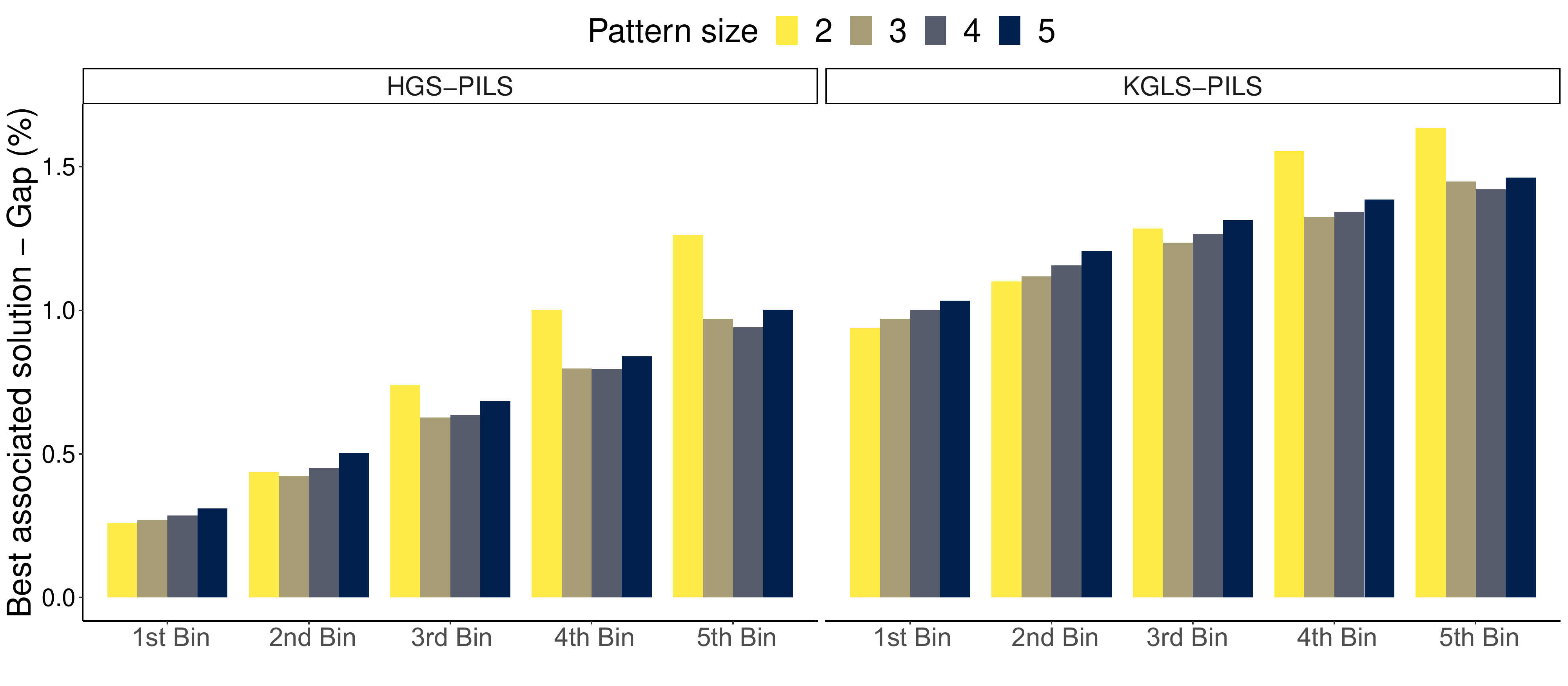}
\caption{Pattern frequency and quality of the associated solutions}
\label{fig:qual}
\end{figure}

Since frequent patterns appear more frequently in optimal solutions, we can also examine whether they are also found in higher-quality solutions in general. We therefore conduct an additional analysis that consist in storing, during the search, each pattern along with the objective value of the best solution in which it appeared. As in the previous experiment, the patterns are sorted by frequency and grouped into equal-sized bins containing $n$ patterns each. Figure~\ref{fig:qual} reports the average quality Gap(\%) of each bin over all runs and instances.

These results confirm the positive correlation between pattern frequency and solution quality. The most frequent patterns are, on average, found in solutions of better quality, up to \myred{0.30\%} better when comparing the solution quality associated with the patterns of the first bin with that of the last bin considered in the figure (fifth bin). These two experiments confirm our initial hypothesis that pattern frequency is a good surrogate for quality, and allow to concentrate the mining procedure on frequent patterns without a need for other filters related to solution quality.

\subsection{Performance impact of PILS}
\label{sec:PILS-quality}

Our last experiment evaluates the performance impact of PILS 
when applied on instances with different characteristics. Table~\ref{tbl:results} displays the results of the comparison of the classical HGS with HGS-PILS and KGLS with KGLS-PILS. For each pair of methods and each instance subset, this table displays the average gap values of both approaches, the percentage time spent in PILS, as well as the result of a paired-samples Wilcoxon test (at a significance level of $p=0.05$) evaluating the statistical significance of the performance difference. The first line corresponds to the complete set of instances, whereas each other line selects a subset of instances with different characteristics, e.g., size, depot location, route length, and customer distribution, using the same nomenclature as in \cite{Uchoa2017}.

\begin{table}[htbp]
\caption{Impact of PILS on solution quality for HGS and KGLS different subsets of instances}\label{tbl:results}
\hspace*{-0.4cm}
\renewcommand{\arraystretch}{1.2}
\setlength{\tabcolsep}{4pt} 
\scalebox{0.92}
{
 \begin{tabular}{lccccccccccccccH} 
 \toprule
 &&\textbf{HGS}&&\multicolumn{2}{c}{\textbf{HGS-PILS}}&&\textbf{Sign.}&&\textbf{KGLS}&&\multicolumn{2}{c}{\textbf{KGLS-PILS}}&&\textbf{Sign.}\\\cmidrule{3-3}\cmidrule{5-6}\cmidrule{8-8}\cmidrule{10-10}\cmidrule{12-13}\cmidrule{15-15}
 Category&\#&Gap(\%)&&Gap(\%)&T$_\textsc{PILS}$(\%)&&&&Gap(\%)&&Gap(\%)&T$_\textsc{PILS}$(\%)&&&T(min)\\
 \midrule
 \rowcolor{Gray} All&100&0.273&&0.242&45.86&&\cmark&\hspace*{0.7cm}&0.555&&0.520&7.28&&\cmark&16.48\\
\midrule
 Smallest&50&0.129&&0.108&45.15&&\cmark&&0.413&&0.379&6.44&&\cmark&8.55\\
 Largest&50&0.418&&0.376&46.56&&\cmark&&0.696&&0.662&8.12&&\cmark&24.36\\
 \midrule
 Short routes&40&0.226&&0.206&44.08&&&&0.581&&0.517&7.64&&\cmark&15.93\\
 Long routes&40&0.337&&0.280&47.94&&\cmark&&0.564&&0.548&6.60&&&17.03\\
 \midrule
 Depot (R)&34&0.317&&0.287&46.45&&&&0.635&&0.558&7.51&&\cmark&17.10\\
 Depot (E)&34&0.267&&0.203&45.15&&\cmark&&0.487&&0.468&5.92&&&16.48\\
 Depot (C)&32&0.234&&0.235&45.99&&&&0.542&&0.537&8.48&&\cmark&15.83\\
 \midrule
 Customer (RC)&34&0.258&&0.228&46.81&&\cmark&&0.553&&0.515&8.13&&\cmark&16.63\\
 Customer (C)&32&0.259&&0.227&45.01&&\cmark&&0.551&&0.545&6.40&&\cmark&16.03\\
 Customer (R)&34&0.301&&0.271&45.70&&\cmark&&0.560&&0.503&7.26&&\cmark &16.75\\
 \bottomrule
 \end{tabular}
}
\end{table}

The results in Table \ref{tbl:results} show that PILS improves the overall performance of both metaheuristics despite their structural differences (population-based versus local search-based). The average gap of HGS over all instances decreases by 0.031\% when combined with PILS, while the average gap of KGLS decreases by 0.035\%. Solution improvements become increasingly difficult as we approach the optimal or best-known values, such then even a small quality improvement of the order of $0.03\%$ over the state-of-the-art is an important achievement.

PILS benefits the search equally on small and large instances, with significant effects observed in both cases. It also improves the performance of HGS for instances with long routes containing many customer visits, likely due to the fact that it compensates for the simplicity of its intra-route neighborhood operators. In contrast, KGLS already uses a sophisticated implementation of Lin-Kernighan algorithm for effective intra-route optimization, but encounters more difficulties to optimize customer allocations among different routes on instances with short routes (i.e., larger customer demands relative to the vehicle capacities). In this situation, we observe that PILS significantly boosts KGLS performance with complementary moves that compensate for this weakness.

\begin{sidewaysfigure}
	\centering
	\begin{minipage}[t]{.49\textwidth}
		\centering
		 \includegraphics[width=\textwidth,height=\textheight,keepaspectratio]{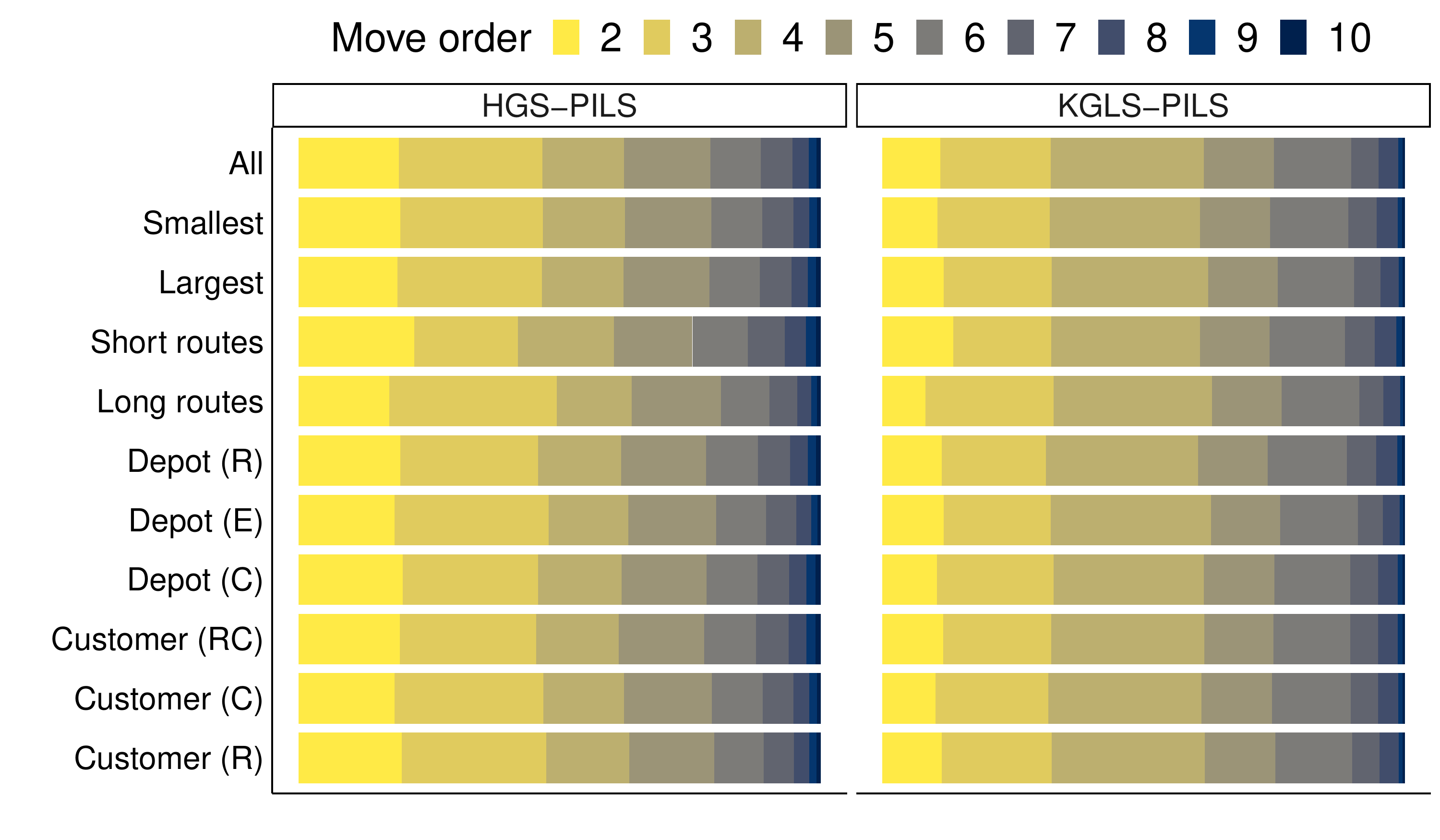}
		\caption{Proportion of applied PILS moves per ``move order''}
		\label{figure-stat1}
	\end{minipage}
	\hfill 
	\begin{minipage}[t]{.49\textwidth}
		\centering
		 \includegraphics[width=\textwidth,height=\textheight,keepaspectratio]{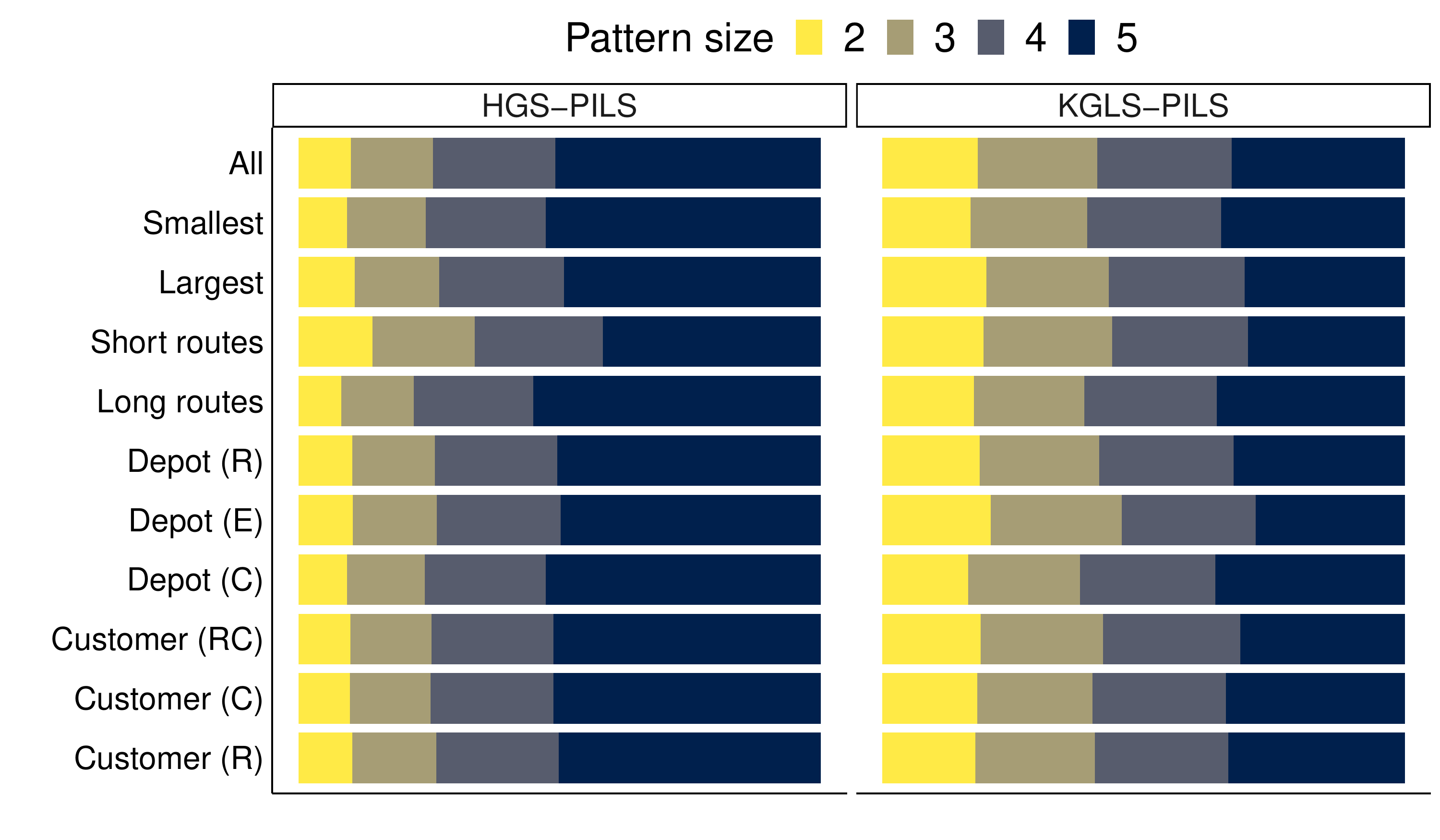}
		\caption{Proportion of applied PILS moves per pattern size}
		\label{figure-stat2}
\end{minipage}
	\vskip\baselineskip
	\begin{minipage}[t]{.49\textwidth}
		\centering
		\includegraphics[width=\textwidth,height=\textheight,keepaspectratio]{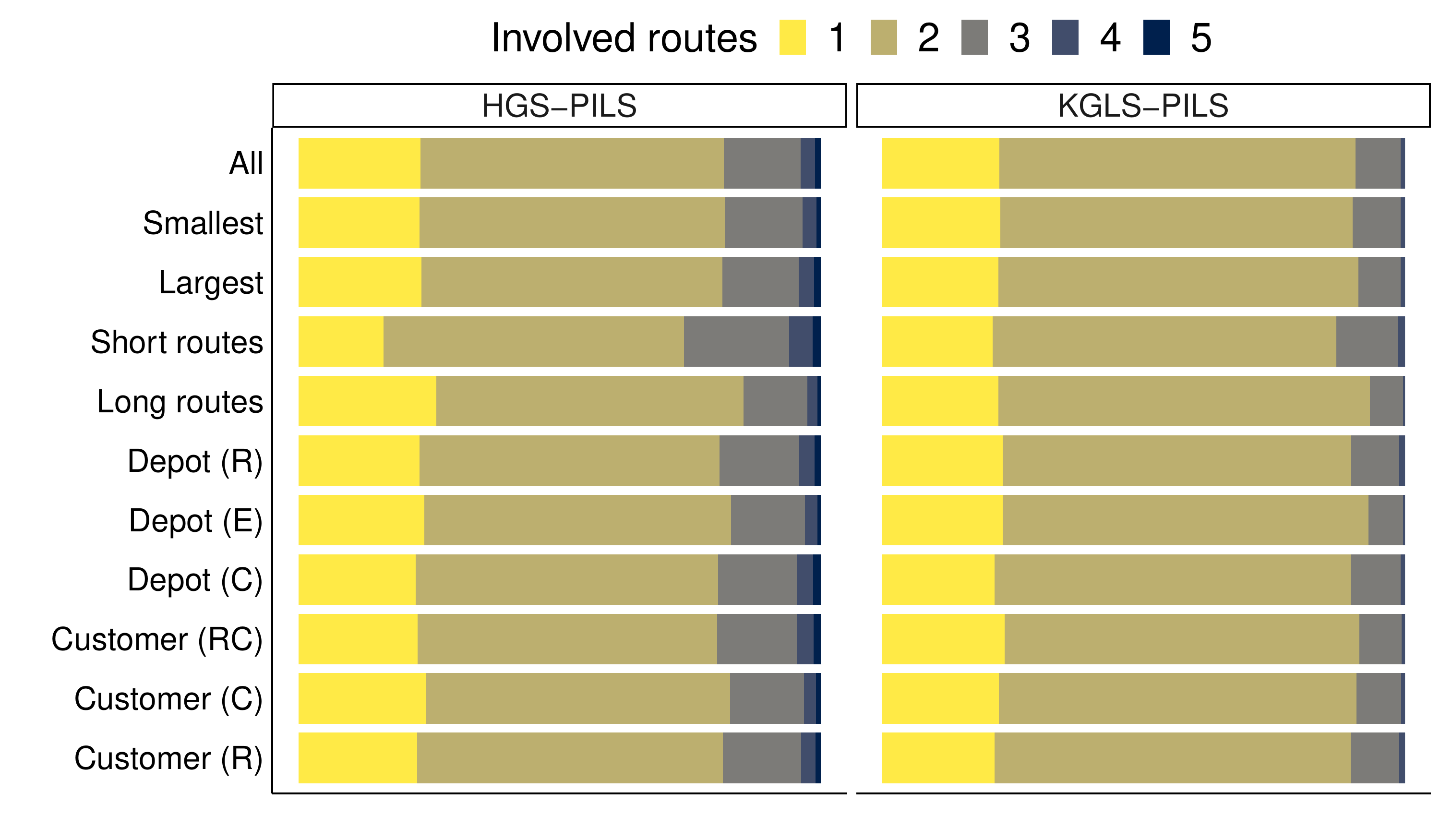}
		\caption{Proportion of applied PILS moves per number of involved routes}
		\label{figure-stat3}
\end{minipage}
		\hfill 
	\begin{minipage}[t]{.49\textwidth}
		\centering
		\includegraphics[width=\textwidth,height=\textheight,keepaspectratio]{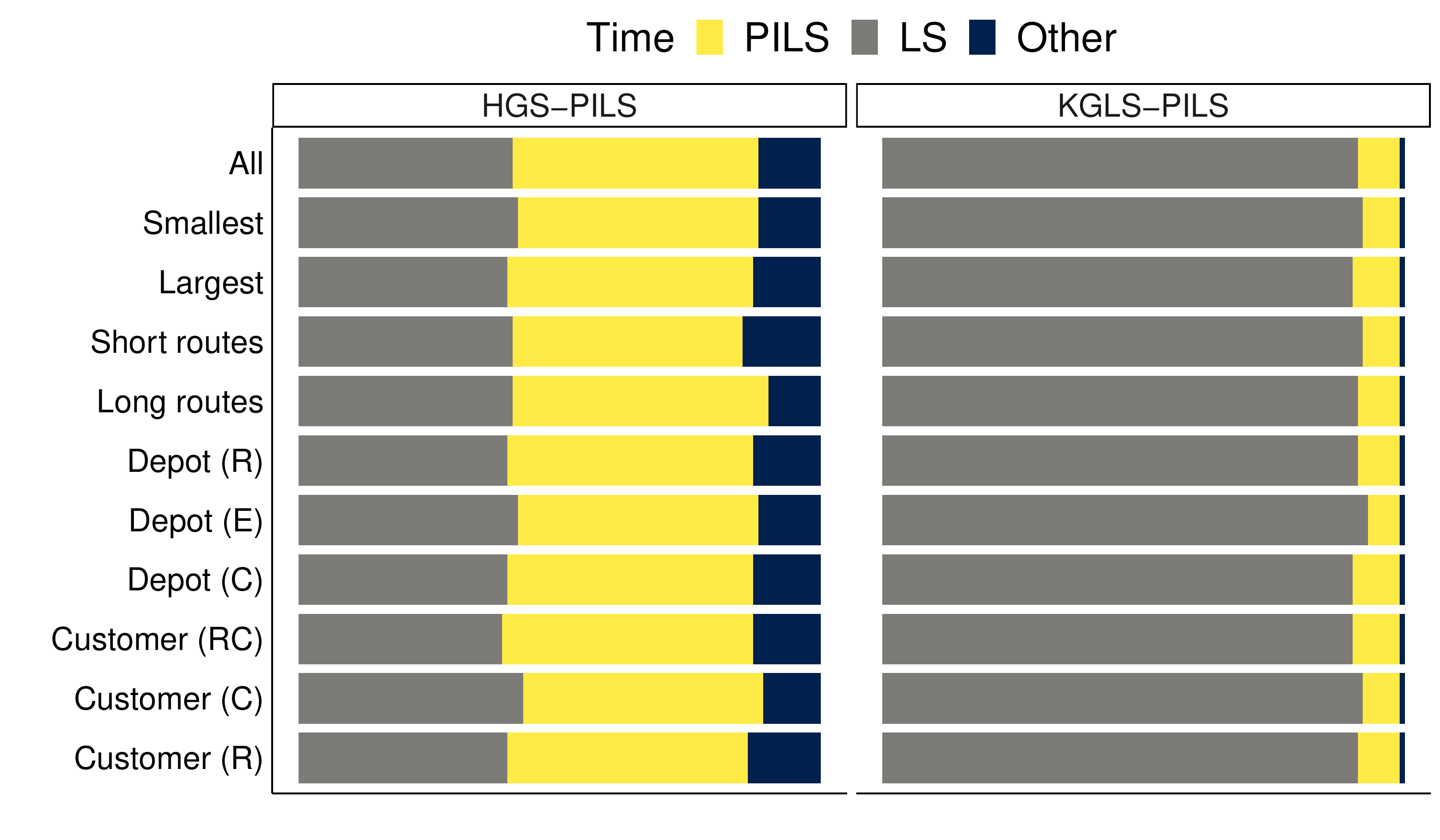}
		\caption{Proportion of CPU time spent in different components of the algorithms}
		\label{figure-stat4}
\end{minipage}
\end{sidewaysfigure}

To gain more insights into the moves that are applied by PILS, we collect a variety of statistics about the injected patterns, as reported in Figures~\ref{figure-stat1} to~\ref{figure-stat4}. These figures represent the proportion of applied PILS moves for each ``move order'' (i.e., number of replaced edges), pattern size, and number of involved routes, during all HGS-PILS and KGLS-PILS executions.

From these experiments, we observe that the largest and smallest patterns are equally likely to be injected. The majority of PILS moves (approximately \myred{80\%}) modify two to five edges, but some larger moves involving up to ten edges are also found and applied to improve the solutions. The ability to find such high-order moves (e.g., improving \textsc{9-opt} or \textsc{10-opt} moves) in a controllable amount of time is noteworthy. Finally, the proportion of time dedicated to PILS remains stable for all subgroups of instances, never exceeding more than 50\% of the total search effort for HGS-PILS, and 10\% of the total search effort for KGLS-PILS.

\begin{figure}[htbp]
 \centering
  \includegraphics[scale=0.92]{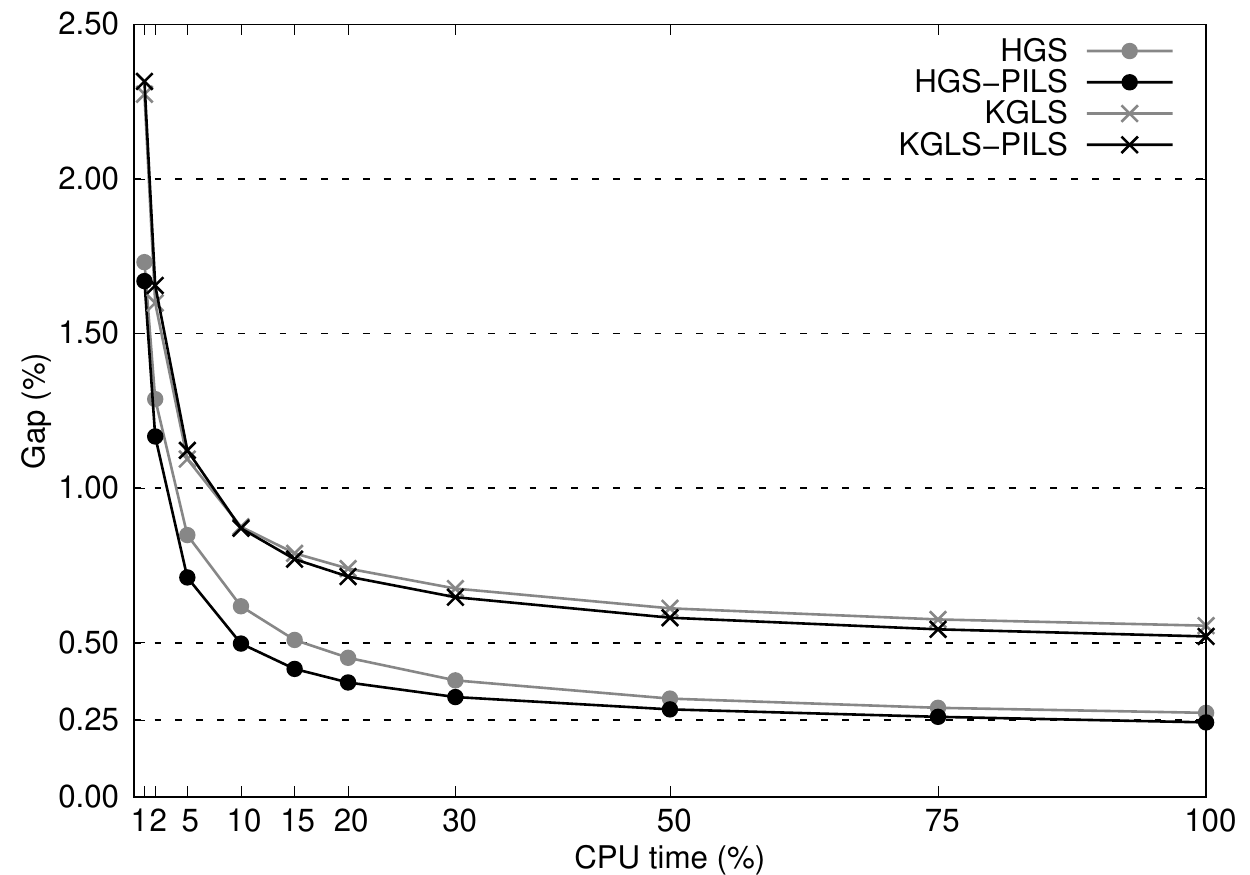}
\caption{Convergence of HGS, HGS-PILS, KGLS and KGLS-PILS solutions over time}
 \label{fig:convergence}
\end{figure}

In a final analysis, Figure \ref{fig:convergence} shows to which extent PILS influences the performance of the two metaheuristics over time. It reports the average Gap(\%) of HGS, HGS-PILS, KGLS and KGLS-PILS at different time steps: after 1\%, 2\%, 5\%, 10\%, 15\%, 20\%, 30\%, 50\%, 75\% and 100\% of the allotted time. One would expect that PILS requires some time to learn good solution patterns and, therefore, that it contributes to the search performance only at later stages. Yet, in the case of HGS-PILS, our experiments do not necessarily corroborate this initial intuition since PILS already boosts the convergence at early stages of the search (e.g., after 10\% of the computational time), leading to solutions of higher quality. In contrast, PILS impacts the search trajectory of KGLS only at later search stages. A likely cause for this observation is that KGLS performs extraction steps only from a single incumbent solution instead of from a population, and thus it requires more time to learn a diversified set of patterns. 

\section{Conclusions}
\label{sec:conclusion}

In this work, we have introduced PILS: a simple and versatile strategy to identify high-order local-search moves using frequent pattern mining.
Our PILS application to the CVRP, a notoriously difficult combinatorial optimization problem, is built upon an effective recursive algorithm that optimally rebuilds solutions from a set of route fragments and a pattern. We integrated PILS into two structurally different state-of-the-art metaheuristics, one population-based algorithm and one trajectory-based algorithm, to evaluate its ability to find new moves and contribute to the search performance. Our experiments confirm that pattern frequency is positively correlated with solution quality and pattern appearance probability within optimal solutions. Moreover, for a fixed computational effort, PILS significantly enhances metaheuristic performance and compensates the weaknesses of each metaheuristic for specific instance subgroups. It complements classical local search operators and identifies synergistic high-order moves (e.g, \textsc{9-opt} or \textsc{10-opt}) which would never be found otherwise.

Numerous possible future research avenues arise from this study.
Firstly, PILS can easily be extended to different combinatorial optimization problems and to solution structures that may require more sophisticated pattern extraction strategies. Secondly, one could also attempt to learn \emph{undesirable} solution patterns to complement the information of the promising ones. Such patterns should of course be \emph{removed} from solutions rather than inserted into them.

Finally, from a more general viewpoint, PILS research occurs in a context in which metaheuristic and pattern recognition research can mutually benefit from each other. Indeed, pattern recognition models often lead to intractable problems which call for efficient heuristic solution approaches, while metaheuristic research directly benefits from enhanced learning strategies.
Indeed, the largest part of metaheuristic research, over the past two decades, has been dedicated to finding simple and efficient strategies to guide surrogate (e.g., constructive or local search) heuristics towards promising search-space regions, construction decisions and moves. Learning desirable solution structures is therefore a defining task for metaheuristic search. Given the tremendous experimental and theoretical progress recently made on a variety of learning algorithms (e.g., belief propagation, deep neural networks, reinforcement learning) and their successful application to some combinatorial optimization problems \citep[e.g.,][]{Bayati2011,Dai2017b}, it is increasingly important to join the strengths and analysis techniques of both fields, to progress towards a new generation of algorithms which remain conceptually simple, amenable to analytic reasoning, and effective. We hope that this first study connecting pattern mining and local search will encourage future work in this direction.

\section*{Acknowledgments}

This research is partially supported by CAPES, CNPq [grant number 308528/2018-2] and FAPERJ [grant number E-26/202.790/2019] in Brazil.
This support is gratefully acknowledged.


\begin{thebibliography}{47}
\expandafter\ifx\csname natexlab\endcsname\relax\def\natexlab#1{#1}\fi
\expandafter\ifx\csname url\endcsname\relax
  \def\url#1{{\tt #1}}\fi
\expandafter\ifx\csname urlprefix\endcsname\relax\def\urlprefix{URL }\fi
\expandafter\ifx\csname urlstyle\endcsname\relax
  \expandafter\ifx\csname doi\endcsname\relax
  \def\doi#1{doi:\discretionary{}{}{}#1}\fi \else
  \expandafter\ifx\csname doi\endcsname\relax
  \def\doi{doi:\discretionary{}{}{}\begingroup \urlstyle{rm}\Url}\fi \fi

\bibitem[{Aggarwal(2014)}]{Aggarwal2014}
Aggarwal, C.C. 2014.
\newblock An introduction to frequent pattern mining.
\newblock C.C. Aggarwal, J.~Han, eds., {\it Frequent Pattern Mining\/}.
  Springer, Cham, 1--17.

\bibitem[{Applegate et~al.(2009)Applegate, Bixby, Chvatal, Cook, Espinoza,
  Goycoolea, and Helsgaun}]{Applegate2009}
Applegate, D., R.~Bixby, V.~Chvatal, W.~Cook, D.~Espinoza, M.~Goycoolea,
  K.~Helsgaun. 2009.
\newblock {Certification of an optimal TSP tour through 85,900 cities}.
\newblock {\it Operations Research Letters\/} {\bf 37}(1) 11--15.

\bibitem[{Arnold et~al.(2019)Arnold, Gendreau, and
  S{\"{o}}rensen}]{Arnold20192}
Arnold, F., M.~Gendreau, K.~S{\"{o}}rensen. 2019.
\newblock {Efficiently solving very large scale routing problems}.
\newblock {\it Computers {\&} Operations Research\/} {\bf 107}(1) 32--42.

\bibitem[{Arnold and S{\"{o}}rensen(2019)}]{Arnold2019}
Arnold, F., K.~S{\"{o}}rensen. 2019.
\newblock {Knowledge-guided local search for the vehicle routing problem}.
\newblock {\it Computers \& Operations Research\/} {\bf 105} 32--46.

\bibitem[{Bahrololoum and Nezamabadi-pour(2017)}]{Bahrololoum2017}
Bahrololoum, A., H.~Nezamabadi-pour. 2017.
\newblock {A multi-expert based framework for automatic image annotation}.
\newblock {\it Pattern Recognition\/} {\bf 61} 169--184.

\bibitem[{Barbalho et~al.(2013)Barbalho, Rosseti, Martins, and
  Plastino}]{Barbalho2013}
Barbalho, H., I.~Rosseti, S.L. Martins, A.~Plastino. 2013.
\newblock {A hybrid data mining GRASP with path-relinking}.
\newblock {\it Computers {\&} Operations Research\/} {\bf 40}(12) 3159--3173.

\bibitem[{Bayati et~al.(2011)Bayati, Borgs, Chayes, and Zecchina}]{Bayati2011}
Bayati, M., C.~Borgs, J.~Chayes, R.~Zecchina. 2011.
\newblock {Belief propagation for weighted b-matchings on arbitrary graphs and
  its relation to linear programs with integer solutions}.
\newblock {\it SIAM Journal on Discrete Mathematics\/} {\bf 25}(2) 989--1011.

\bibitem[{Blum and Roli(2003)}]{Blum2003}
Blum, C., A.~Roli. 2003.
\newblock {Metaheuristics in combinatorial optimization: Overview and
  conceptual comparison}.
\newblock {\it ACM Computing Surveys\/} {\bf 35}(3) 268--308.

\bibitem[{Boese(1995)}]{Boese1995}
Boese, K.D. 1995.
\newblock {Cost versus distance in the traveling salesman problem}.
\newblock Tech. rep., UCLA Computer Science Dept, Los Angeles.

\bibitem[{Christiaens and {Vanden Berghe}(2018)}]{Christiaens2018}
Christiaens, J., G.~{Vanden Berghe}. 2018.
\newblock Slack induction by string removals for vehicle routing problems.
\newblock Tech. rep., KU Leuven, Department of Computer Science, CODeS \& imec,
  Leuven.

\bibitem[{Costa et~al.(2019)Costa, Contardo, and Desaulniers}]{Costa2019}
Costa, L., C.~Contardo, G.~Desaulniers. 2019.
\newblock {Exact branch-price-and-cut algorithms for vehicle routing}.
\newblock {\it Transportation Science\/} {\bf 53}(4) 946--985.

\bibitem[{Dai et~al.(2017)Dai, Khalil, Zhang, Dilkina, and Song}]{Dai2017b}
Dai, H., E.B. Khalil, Y.~Zhang, B.~Dilkina, L.~Song. 2017.
\newblock {Learning combinatorial optimization algorithms over graphs}.
\newblock {\it Advances in Neural Information Processing Systems\/}
  6348--6358.

\bibitem[{Dorigo and St{\"u}tzle(2019)}]{Dorigo2019}
Dorigo, M., T.~St{\"u}tzle. 2019.
\newblock Ant colony optimization: Overview and recent advances.
\newblock M.~Gendreau, J.-Y. Potvin, eds., {\it Handbook of Metaheuristics\/},
  3rd ed. Springer, Boston, 311--351.

\bibitem[{{El Hachemi} et~al.(2015){El Hachemi}, Crainic, Lahrichi, Rei, and
  Vidal}]{ElHachemi2015}
{El Hachemi}, N., T.G. Crainic, N.~Lahrichi, W.~Rei, T.~Vidal. 2015.
\newblock {Solution integration in combinatorial optimization with applications
  to cooperative search and rich vehicle routing}.
\newblock {\it Journal of Heuristics\/} {\bf 21}(5) 663--685.

\bibitem[{Gendreau et~al.(1994)Gendreau, Hertz, and Laporte}]{Gendreau1994}
Gendreau, M., A.~Hertz, G.~Laporte. 1994.
\newblock {A tabu search heuristic for the vehicle routing problem}.
\newblock {\it Management Science\/} {\bf 40}(10) 1276--1290.

\bibitem[{Glover(1997)}]{Glover1997}
Glover, F. 1997.
\newblock {Tabu search and adaptive memory programming -- Advances,
  applications and challenges}.
\newblock R.S. Barr, R.V. Helgason, J.L. Kennington, eds., {\it Advances in
  Metaheuristics, Optimization, and Stochastic Modeling Technologies\/}.
  Springer, Boston, 1--75.

\bibitem[{Gribel and Vidal(2019)}]{Gribel2019}
Gribel, D., T.~Vidal. 2019.
\newblock {HG-means: A scalable hybrid metaheuristic for minimum sum-of-squares
  clustering}.
\newblock {\it Pattern Recognition\/} {\bf 88} 569--583.

\bibitem[{Han et~al.(2019)Han, Jiang, Ling, and Su}]{Han2019a}
Han, F., J.~Jiang, Q.-H. Ling, B.Y. Su. 2019.
\newblock {A survey on metaheuristic optimization for random single-hidden
  layer feedforward neural network}.
\newblock {\it Neurocomputing\/} {\bf 335} 261--273.

\bibitem[{Hansen et~al.(2012)Hansen, Ruiz, and Aloise}]{Hansen2012}
Hansen, P., M.~Ruiz, D.~Aloise. 2012.
\newblock {A VNS heuristic for escaping local extrema entrapment in normalized
  cut clustering}.
\newblock {\it Pattern Recognition\/} {\bf 45}(12) 4337--4345.

\bibitem[{Holland(1992)}]{Holland1992}
Holland, J.H. 1992.
\newblock {\it {Adaptation in natural and artificial systems: An introductory
  analysis with applications to biology, control, and artificial
  intelligence}\/}.
\newblock MIT Press, Cambridge.

\bibitem[{Ijjina and Chalavadi(2016)}]{Ijjina2016}
Ijjina, E.P., K.M. Chalavadi. 2016.
\newblock {Human action recognition using genetic algorithms and convolutional
  neural networks}.
\newblock {\it Pattern Recognition\/} {\bf 59} 199--212.

\bibitem[{Jovanovic et~al.(2019)Jovanovic, Tuba, and Vo{\ss}}]{Jovanovic2019}
Jovanovic, R., M.~Tuba, S.~Vo{\ss}. 2019.
\newblock Fixed set search applied to the traveling salesman problem.
\newblock M.J.B. Aguilera, C.~Blum, H.G. Santos, P.~Pinacho, J.~Godoy, eds.,
  {\it Hybrid Metaheuristics\/}, {\it Lecture Notes in Computer Science\/},
  vol. 11299. Springer, Cham, 63--77.

\bibitem[{Kashef and Nezamabadi-pour(2015)}]{Kashef2015a}
Kashef, S., H.~Nezamabadi-pour. 2015.
\newblock {An advanced ACO algorithm for feature subset selection}.
\newblock {\it Neurocomputing\/} {\bf 147}(1) 271--279.

\bibitem[{Kilby et~al.(2005)Kilby, Slaney, and Walsh}]{Kilby2005}
Kilby, P., J.~Slaney, T.~Walsh. 2005.
\newblock The backbone of the travelling salesperson.
\newblock {\it Proceedings of the 19th International Joint Conference on
  Artificial Intelligence\/}. San Francisco, CA, USA, 175--180.

\bibitem[{Lahrichi et~al.(2015)Lahrichi, Crainic, Gendreau, Rei, Crişan, and
  Vidal}]{Lahrichi2015}
Lahrichi, N., T.G. Crainic, M.~Gendreau, W.~Rei, G.C. Crişan, T.~Vidal. 2015.
\newblock {An integrative cooperative search framework for
  multi-decision-attribute combinatorial optimization: Application to the
  MDPVRP}.
\newblock {\it European Journal of Operational Research\/} {\bf 246}(2)
  400--412.

\bibitem[{{Le Bouthillier} et~al.(2005){Le Bouthillier}, Crainic, and
  Kropf}]{LeBouthillier2005}
{Le Bouthillier}, A., T.G. Crainic, P.~Kropf. 2005.
\newblock {A guided cooperative search for the vehicle routing problem with
  time windows}.
\newblock {\it IEEE Intelligent Systems\/} {\bf 20}(4) 36--42.

\bibitem[{Lin and Kernighan(1973)}]{lin1973effective}
Lin, S., B.W. Kernighan. 1973.
\newblock An effective heuristic algorithm for the traveling-salesman problem.
\newblock {\it Operations Research\/} {\bf 21}(2) 498--516.

\bibitem[{Muter et~al.(2010)Muter, Birbil, and Şahin}]{Muter2010}
Muter, I., Ş.I. Birbil, G.~Şahin. 2010.
\newblock {Combination of metaheuristic and exact algorithms for solving set
  covering-type optimization problems}.
\newblock {\it INFORMS Journal on Computing\/} {\bf 22}(4) 603--619.

\bibitem[{Nagata and Br{\"{a}}ysy(2009)}]{Nagata2009}
Nagata, Y., O.~Br{\"{a}}ysy. 2009.
\newblock Edge assembly-based memetic algorithm for the capacitated vehicle
  routing problem.
\newblock {\it Networks\/} {\bf 54}(4) 205--215.

\bibitem[{Pecin et~al.(2017)Pecin, Pessoa, Poggi, and Uchoa}]{Pecin2017}
Pecin, D., A.~Pessoa, M.~Poggi, E.~Uchoa. 2017.
\newblock Improved branch-cut-and-price for capacitated vehicle routing.
\newblock {\it Mathematical Programming Computation\/} {\bf 9}(1) 61--100.

\bibitem[{Prins(2004)}]{Prins2004}
Prins, C. 2004.
\newblock A simple and effective evolutionary algorithm for the vehicle routing
  problem.
\newblock {\it Computers \& Operations Research\/} {\bf 31}(12) 1985--2002.

\bibitem[{Resende et~al.(2010)Resende, Ribeiro, Glover, and
  Mart{\'{i}}}]{Resende2010}
Resende, M.G.C., C.C. Ribeiro, F.~Glover, R.~Mart{\'{i}}. 2010.
\newblock {Scatter search and path-relinking: Fundamentals, advances, and
  applications}.
\newblock M.~Gendreau, J.-Y. Potvin, eds., {\it Handbook of Metaheuristics\/},
  2nd ed. Springer, Boston, 87--107.

\bibitem[{Ribeiro et~al.(2006)Ribeiro, Plastino, and Martins}]{Ribeiro2006}
Ribeiro, M.H., A.~Plastino, S.L. Martins. 2006.
\newblock {Hybridization of GRASP metaheuristic with data mining techniques}.
\newblock {\it Journal of Mathematical Modelling and Algorithms\/} {\bf 5}(1)
  23--41.

\bibitem[{Santos et~al.(2006)Santos, Ochi, Marinho, and Drummond}]{Santos2006}
Santos, H.G., L.S. Ochi, E.H. Marinho, L.M.A. Drummond. 2006.
\newblock {Combining an evolutionary algorithm with data mining to solve a
  single-vehicle routing problem}.
\newblock {\it Neurocomputing\/} {\bf 70}(1-3) 70--77.

\bibitem[{Schneider(2003)}]{Schneider2003}
Schneider, J. 2003.
\newblock Searching for backbones -- a high-performance parallel algorithm for
  solving combinatorial optimization problems.
\newblock {\it Future Generation Computer Systems\/} {\bf 19}(1) 121--131.

\bibitem[{Subramanian et~al.(2013)Subramanian, Uchoa, and
  Ochi}]{Subramanian2013}
Subramanian, A., E.~Uchoa, L.S. Ochi. 2013.
\newblock {A hybrid algorithm for a class of vehicle routing problems}.
\newblock {\it Computers {\&} Operations Research\/} {\bf 40}(10) 2519--2531.

\bibitem[{Taillard et~al.(2001)Taillard, Gambardella, Gendreau, and
  Potvin}]{Taillard2001}
Taillard, {\'{E}}.D., L.M. Gambardella, M.~Gendreau, J.-Y. Potvin. 2001.
\newblock {Adaptive memory programming: A unified view of metaheuristics}.
\newblock {\it European Journal of Operational Research\/} {\bf 135}(1) 1--16.

\bibitem[{Tarantilis and Kiranoudis(2002)}]{Tarantilis2002}
Tarantilis, C.D., C.T. Kiranoudis. 2002.
\newblock {BoneRoute: An adaptive memory-based method for effective fleet
  management}.
\newblock {\it Annals of Operations Research\/} {\bf 115}(1--4) 227--241.

\bibitem[{Toth and Vigo(2014)}]{Toth2014}
Toth, P., D.~Vigo. 2014.
\newblock {\it {Vehicle routing: Problems, methods, and applications}\/}.
\newblock 2nd ed. MOS-SIAM Series on Optimization, Philadelphia.

\bibitem[{Uchoa et~al.(2017)Uchoa, Pecin, Pessoa, Poggi, Vidal, and
  Subramanian}]{Uchoa2017}
Uchoa, E., D.~Pecin, A.~Pessoa, M.~Poggi, T.~Vidal, A.~Subramanian. 2017.
\newblock New benchmark instances for the capacitated vehicle routing problem.
\newblock {\it European Journal of Operational Research\/} {\bf 257}(3)
  845--858.

\bibitem[{Vidal et~al.(2012)Vidal, Crainic, Gendreau, Lahrichi, and
  Rei}]{Vidal2012}
Vidal, T., T.G. Crainic, M.~Gendreau, N.~Lahrichi, W.~Rei. 2012.
\newblock {A hybrid genetic algorithm for multidepot and periodic vehicle
  routing problems}.
\newblock {\it Operations Research\/} {\bf 60}(3) 611--624.

\bibitem[{Vidal et~al.(2013)Vidal, Crainic, Gendreau, and Prins}]{Vidal2012a}
Vidal, T., T.G. Crainic, M.~Gendreau, C.~Prins. 2013.
\newblock {Heuristics for multi-attribute vehicle routing problems: A survey
  and synthesis}.
\newblock {\it European Journal of Operational Research\/} {\bf 231}(1) 1--21.

\bibitem[{Vidal et~al.(2014)Vidal, Crainic, Gendreau, and Prins}]{Vidal2014}
Vidal, T., T.G. Crainic, M.~Gendreau, C.~Prins. 2014.
\newblock {A unified solution framework for multi-attribute vehicle routing
  problems}.
\newblock {\it European Journal of Operational Research\/} {\bf 234}(3)
  658--673.

\bibitem[{Vidal et~al.(2019)Vidal, Laporte, and Matl}]{Vidal2019}
Vidal, T., G.~Laporte, P.~Matl. 2019.
\newblock {A concise guide to existing and emerging vehicle routing problem
  variants}.
\newblock {\it European Journal of Operational Research, Articles in Advance\/}
  .

\bibitem[{Voudouris and Tsang(2003)}]{voudouris2003guided}
Voudouris, Christos, Edward~PK Tsang. 2003.
\newblock {\it Guided local search\/}.
\newblock Springer.

\bibitem[{Wolpert(1997)}]{Wolpert1997}
Wolpert, D.H. 1997.
\newblock {No free lunch theorems for optimization}.
\newblock {\it IEEE Transactions on Evolutionary Computation\/} {\bf 1}(1)
  67--82.

\bibitem[{Yusta(2009)}]{Yusta2009}
Yusta, S.C. 2009.
\newblock {Different metaheuristic strategies to solve the feature selection
  problem}.
\newblock {\it Pattern Recognition Letters\/} {\bf 30}(5) 525--534.

\end{thebibliography}

\end{document}